%% file: main.tex
\documentclass{article} 
\PassOptionsToPackage{table}{xcolor}
\usepackage{iclr2027_conference,times}
\usepackage{xcolor}

\input{math_commands.tex}

\usepackage{amssymb}

\usepackage{hyperref}
\usepackage{url}
\usepackage{url}
\usepackage{graphicx}
\usepackage{wrapfig}
\usepackage{booktabs,tabularx}

\title{What Shared Prefixes Hide: Trajectory Dropout for On-Policy Distillation  }

\author{Zizhuo Lin$^{\heartsuit}$, Quanling Liu$^{\heartsuit}$ ,Yi Yang , Yawei Luo$^{\diamondsuit}$\\
Zhejiang University\\
\texttt{\{zizhuolin, liuquanling, yaweiluo\}@zju.edu.cn}%
}

\iclrfinalcopy 
\begin{document}

\maketitle

\renewcommand{\thefootnote}{\ensuremath{\heartsuit}}%
\footnotetext{These authors contributed equally to this work.}%
\setcounter{footnote}{1}%
\renewcommand{\thefootnote}{\ensuremath{\diamondsuit}}%
\footnotetext{Corresponding author.}%
\setcounter{footnote}{0}%

\begin{figure}[!htbp]
 \centering
 \includegraphics[width=0.85\linewidth]{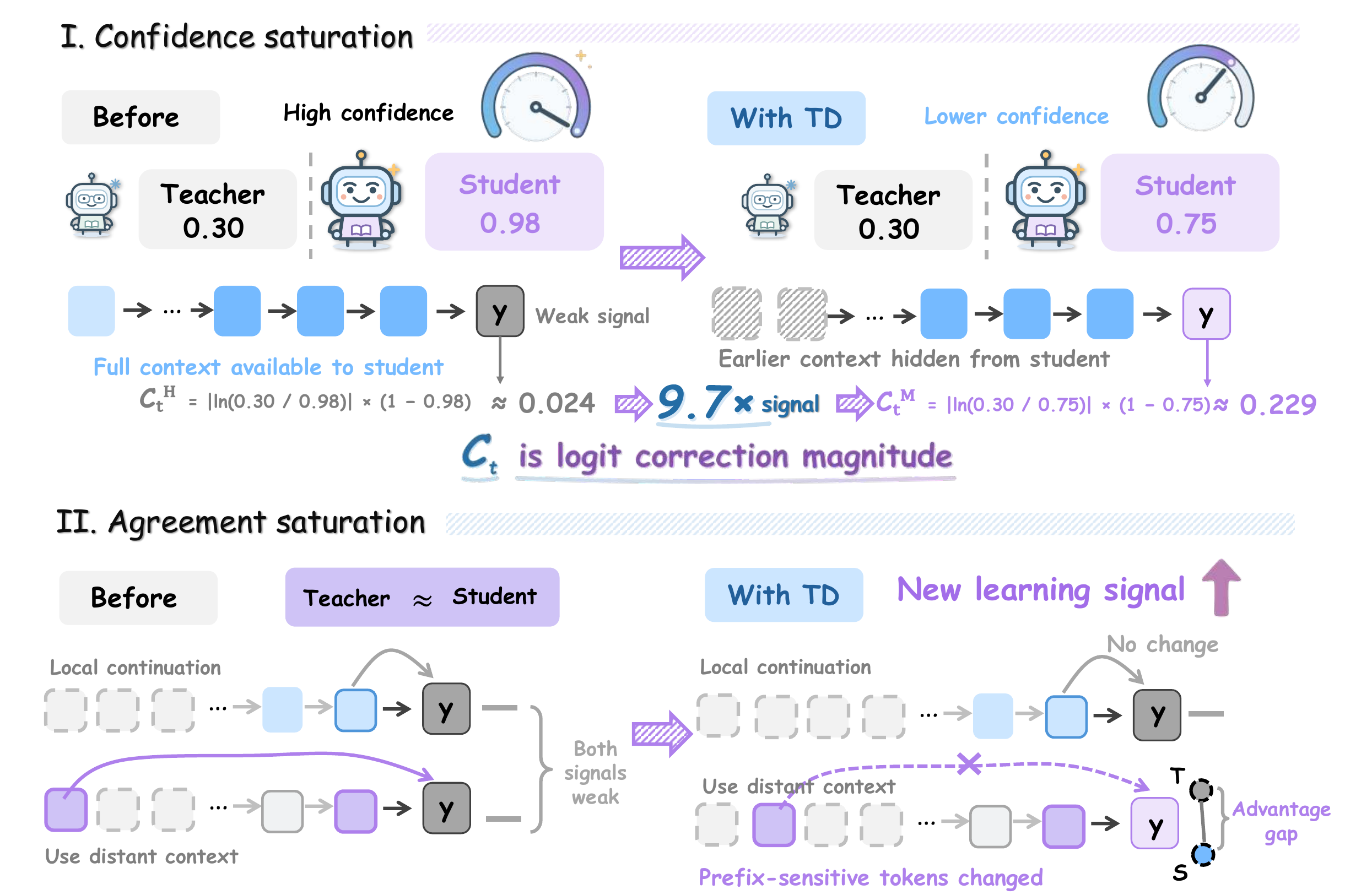}
 \caption{TD-OPD eases confidence and agreement saturation by adding Trajectory Dropout during training, boosts the training signals.
 }
 \label{fig:teaser}
\end{figure}

\begin{abstract}
\input{Section/00-abstruct}
\end{abstract}

\input{Section/01-Introduction}

\input{Section/02-PISA}
\input{Section/03-TD-OPD}

\input{Section/03-Experiment}

\input{Section/Related_work}
\input{Section/04-Conclustion}

\input{Section/05-statement}
\bibliography{iclr2027_conference}
\bibliographystyle{iclr2027_conference}

\appendix

\input{Appendix/other_results}

\input{Appendix/experimental-configurations}
\input{Appendix/evaluation_and_dataset}
\input{Appendix/signal_analysis}
\input{Appendix/Norm-Controlled-Comparison}
\input{Appendix/appendix}

\end{document}

%% file: math_commands.tex
\usepackage{amsmath,amsfonts,bm}

\newcommand{\sg}{\operatorname{sg}}

\def\eqref#1{equation~\ref{#1}}

\def\1{\bm{1}}

\DeclareMathAlphabet{\mathsfit}{\encodingdefault}{\sfdefault}{m}{sl}
\SetMathAlphabet{\mathsfit}{bold}{\encodingdefault}{\sfdefault}{bx}{n}



%% file: Section/00-abstruct.tex
On-policy distillation (OPD) trains a student model on its own trajectories using dense token-level feedback from a stronger teacher model.
Since each update is conditioned on the reasoning prefix already generated by the student, the prefix also shapes how effectively teacher feedback is converted into learning.
We find that shared prefixes can lead to weak token-level updates, a phenomenon we call Prefix-Induced Supervision Attenuation (PISA).
This attenuation arises in two common cases.
(i) High student confidence can weaken corrective gradients even when the teacher disagrees.
(ii) Tokens that rely on earlier reasoning can receive learning signals as weak as those for simple local continuations.
To solve this problem, we propose Trajectory Dropout, a simple training-time intervention that exposes these weakened signals.
The student first performs a standard full-context rollout to generate a complete trajectory.
During training, we randomly drop a certain proportion of the student's reasoning trajectory, while the teacher continues to observe the complete trajectory for token-level supervision.
This intervention strengthens corrections for overconfident predictions and introduces additional supervision at prefix-sensitive positions.
Trajectory Dropout consistently improves average performance across teacher--student model pairs of different scales and six mathematical reasoning benchmarks, while also yielding gains on two out-of-domain benchmarks.
It can also be flexibly integrated into existing OPD variants with negligible computational overhead, further improving their performance.
These results demonstrate that Trajectory Dropout provides a simple mechanism for strengthening token-level supervision across model scales and OPD objectives.

%% file: Section/01-Introduction.tex
\section{Introduction}
On-policy distillation (OPD) trains a student on its own generated
trajectories using dense token-level feedback from a stronger
teacher~\citep{gu2024minillm,agarwal2024gkd}.
This brings supervision to the prefixes encountered during student
generation and reduces the mismatch between training and inference.
Recent studies show that the resulting signals vary along a trajectory.
Training signals can be concentrated near the beginning of a response
~\citep{fastopd2026}.
As student-generated prefixes grow, the teacher's predictions can also
become less discriminative, weakening its corrective
feedback~\citep{liu2026fidelity}.
Errors in the prefix can further pull teacher predictions between
continuing a flawed path and correcting it~\citep{trd2026}.
These findings show that dense feedback can vary substantially in
reliability and strength~\citep{2026dense}.

The learning value of this feedback also differs across tokens.
\citet{tip2026} identify useful signals at both high-entropy positions
and low-entropy positions with strong teacher--student disagreement.
\citet{taopd2026} further show that the value of disagreement
depends on whether the teacher's correction lies within the student's
plausible candidate set.
\cite{helm-etal-2025-token} and \cite{deng2026beyond} show that  tokens whose predictions benefit from non-local context are particularly useful training objectives.
Gradient-based analysis likewise reveals large differences in how
teacher guidance supports learning across token
positions~\citep{armandpour2026unmasking,shao2026token}.
These studies motivate a closer examination of the student's actual
response to dense supervision:

\emph{OPD provides dense token-level supervision—but does dense supervision necessarily translate into dense learning signals?}

We address this question by examining per-token gradient contributions in sampled-token OPD. In our diagnostic setting, approximately 70\% of supervised tokens contribute less than 0.5\% of the measured own-logit signal mass. Teacher feedback covers the trajectory, yet the resulting updates are highly concentrated. Importantly, a weak update does not necessarily imply a low-value training example. Some weakly updated positions correspond to predictions that the teacher still disfavors, while others arise when the complete reasoning history makes qualitatively different predictions appear equally easy. This motivates us to examine how the shared prefix shapes the learning signal received by each token.


We identify \emph{Prefix-Induced Supervision Attenuation} (PISA): useful learning opportunities can be weakly expressed in the full-view update because of the prefix available to the student. PISA appears in two forms. \emph{Confidence saturation} occurs when the prefix supports a highly confident student prediction, making the student insensitive to teacher correction even when disagreement remains. \emph{Agreement saturation} occurs when near-zero teacher--student advantage conflates routine local continuations with predictions that rely on earlier reasoning. The latter are qualitatively different from routine local continuations because their predictability depends on earlier reasoning, yet the full-view advantage does not distinguish between these two cases. Controlled prefix interventions reveal that a substantial subset of these weak-signal positions is sensitive to access to earlier reasoning.


Controlled interventions suggest that the student’s access to its own reasoning history can serve as a useful training-time control. We introduce \emph{Trajectory Dropout} (TD), which selectively restricts this access during student updates while leaving the on-policy rollout and the teacher’s full-prefix supervision unchanged. The student first generates a complete trajectory with standard full-context decoding. During training, TD modifies the student’s access to selected parts of the saved reasoning history, while all target tokens remain supervised by the teacher on their complete causal prefixes.


TD acts on the two weak-signal regimes in complementary ways. For overconfident predictions, reducing prefix support can increase the student’s sensitivity and strengthen an existing teacher-directed correction. For near-agreement predictions, changing prefix access introduces a sensitivity signal that differentiates positions that appear indistinguishable under the complete history. Prefix-sensitive positions can therefore receive additional training signal, while preserving nearby access limits disruption to local predictive dependencies.


We evaluate TD on six mathematical reasoning benchmarks, and two out-of-domain benchmarks.
TD consistently improves average performance over standard OPD under
normal full-prefix inference.
The gains extend to teacher--student pairs at different scales and to
existing OPD variants, including TRD~\citep{trd2026} and
Fast-OPD~\citep{fastopd2026}.
Paired signal analysis supports stronger teacher-directed corrections in the high-confidence regime and additional training signal among prefix-sensitive near-agreement positions.

We further study the conditions that make this prefix intervention
stable.
Ablations over the dropout budget and protection window show that
widely scattered masking can disrupt nearby predictive dependencies.
Allowing positions in a short window after each masked segment to
retain access to that segment stabilizes training.
Protection-window ablations show that preserving nearby predictive access is important when masking becomes highly fragmented.
Together, the results show that controlled prefix perturbation can strengthen or introduce training signals at weakly updated positions while preserving the local predictive structure needed for effective optimization.

Our contributions are threefold:
\begin{itemize}
    \item We identify \textbf{Prefix-Induced Supervision Attenuation (PISA)} in
    sampled-token OPD.
    Gradient analysis and controlled prefix interventions reveal
    learning opportunities among weakly updated tokens.

    \item We introduce \textbf{Trajectory Dropout (TD)}, which preserves standard
    rollouts and full-prefix teacher supervision while changing the
    student's update view.
    It strengthens existing corrections and introduces additional signal that disambiguates prefix-sensitive positions within the near-agreement region.

    \item We demonstrate gains across mathematical reasoning benchmarks,
    model scales, and OPD variants. Ablations support the
    roles of update allocation and local prefix continuity.
\end{itemize}

%% file: Section/02-PISA.tex
\providecommand{\sg}
{\operatorname{sg}}
\providecommand{\FloatBarrier}{}
\providecommand{\Needspace}[1]{}

\section{Prefix-Induced Supervision Attenuation}
\label{sec:method}

\subsection{Token-Level Learning Signals in On-Policy Distillation}
\label{sec:opd}

On-policy distillation (OPD) trains a student on its own responses
\citep{agarwal2024gkd,gu2024minillm}.
The teacher scores each generated token on the same causal prefix.
We analyze the sampled-token OPD update on a fixed rollout batch.
For prompt $x$ and response $y$, write $h_t=(x,y_{<t})$.
Let $p_t=P_T(y_t\mid h_t)$ and $q_{\theta,t}=Q_\theta(y_t\mid h_t)$
be the teacher and student probabilities of the sampled token $y_t$.
The student weights before the update are $\bar\theta$, and $q_t=q_{\bar\theta,t}$.
The raw surrogate uses sampled reverse-KL feedback:
\begin{equation}
 \mathcal L_{\mathrm{OPD}}(\theta)
 =-\frac1N\sum_t\frac{q_{\theta,t}}{q_t}
 \sg\!\left[\log p_t-\log q_{\theta,t}\right].
 \label{eq:opd-loss}
\end{equation}
Here $N$ counts response tokens, and $\sg$ stops gradients through its argument.
The log-probability gap is the token's \emph{advantage}.
A positive advantage favors a higher token probability; a negative one favors a lower probability.

At $\theta=\bar\theta$, the token-level update signal has the product form
\begin{equation}
 A_t^H=\log p_t-\log q_t,\qquad
 u_t^H=A_t^Hs_t^H,\qquad
 s_t^H=\left.\nabla_\theta\log q_{\theta,t}\right|_{\bar\theta}.
 \label{eq:opd-product}
\end{equation}
The superscript $H$ denotes full context.
The advantage measures teacher--student disagreement on $y_t$.
The score vector $s_t^H$ measures the sensitivity of its log probability to the weights.
Their product gives the token's contribution before optimizer transformations.
Either factor can make this contribution weak.
We compare tokens using the output-logit signal
$G_{z,t}^H=|A_t^H|\|\nabla_{z_t^H}\log q_t\|_2$, where $z_t^H$ is the full-view logit vector.
This measures the combined effect of the advantage and output sensitivity.
\emph{Signal mass} is the sum of these token-level norms.
Figure~\ref{fig:td-pisa}A shows its uneven distribution in a full-context audit.

\subsection{Confidence and Agreement Saturation}
\label{sec:pisa}
\begin{figure}[!htbp]
 \centering
 \includegraphics[width=0.9\linewidth]{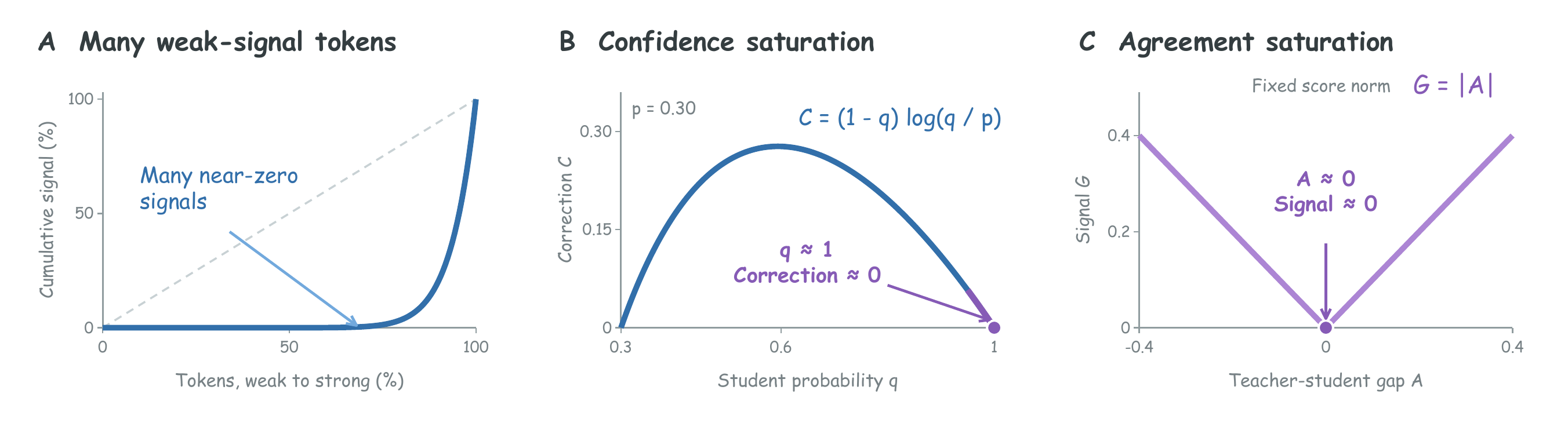}
 \caption{Two routes to weak OPD signals.
 A: cumulative signal magnitude after sorting tokens from weakest to strongest.
 B: the own-logit correction vanishes as student confidence approaches one, for fixed teacher probability.
 C: the signal vanishes near zero advantage, for fixed score norm.
 }
 \label{fig:td-pisa}
\end{figure}

The shared prefix can lead to weak token-level OPD updates through two distinct mechanisms. It can make the student highly confident in a prediction, reducing its sensitivity to teacher correction. It can also make qualitatively different predictions appear equally easy under the full history, collapsing prefix-dependent reasoning steps and routine local continuations into similarly weak teacher–student advantages. We refer to these effects as \emph{Prefix-Induced Supervision Attenuation} (PISA).

\textbf{Confidence Saturation.}
Consider a token with $p_t<q_t\approx1$.
The teacher assigns it a lower probability.
For softmax logits, the student's own-token derivative is $1-q_t$.
The magnitude of the downward signal at this logit is therefore
\begin{equation}
 C_t^H=|A_t^H|(1-q_t)
      =(1-q_t)\log(q_t/p_t).
 \label{eq:confidence}
\end{equation}
For fixed $p_t>0$, this correction approaches zero as $q_t\to1$.
Near $q_t=1$, rising confidence weakens the correction
(Figure~\ref{fig:td-pisa}B).
Appendix~\ref{app:confidence} extends the result to the full logit vector
and gives the corresponding parameter-gradient bound.

\textbf{Agreement Saturation.}
When the two models assign similar log probabilities to $y_t$, $A_t^H\approx0$.
The product in~(\ref{eq:opd-product}) then gives a weak update for a bounded score vector (Figure~\ref{fig:td-pisa}C). Such agreement is not inherently uninformative: it can arise for qualitatively different reasons. Some tokens are routine local continuations that are easy to predict from nearby context, whereas others are easy because they correctly rely on results established earlier in the reasoning trajectory. Under the complete history, both cases can produce similarly small advantages. However, prefix-dependent predictions provide useful training examples for learning how information is carried across reasoning steps and should not be conflated with simple local continuations. The problem is therefore not agreement itself, but that full-view agreement obscures the source of predictability. As a result, valuable learning opportunities at prefix-dependent positions can be buried among routine local continuations with equally weak signals.
\begin{wrapfigure}[15]{r}{0.5\textwidth}
    \centering
    \includegraphics[width=\linewidth]{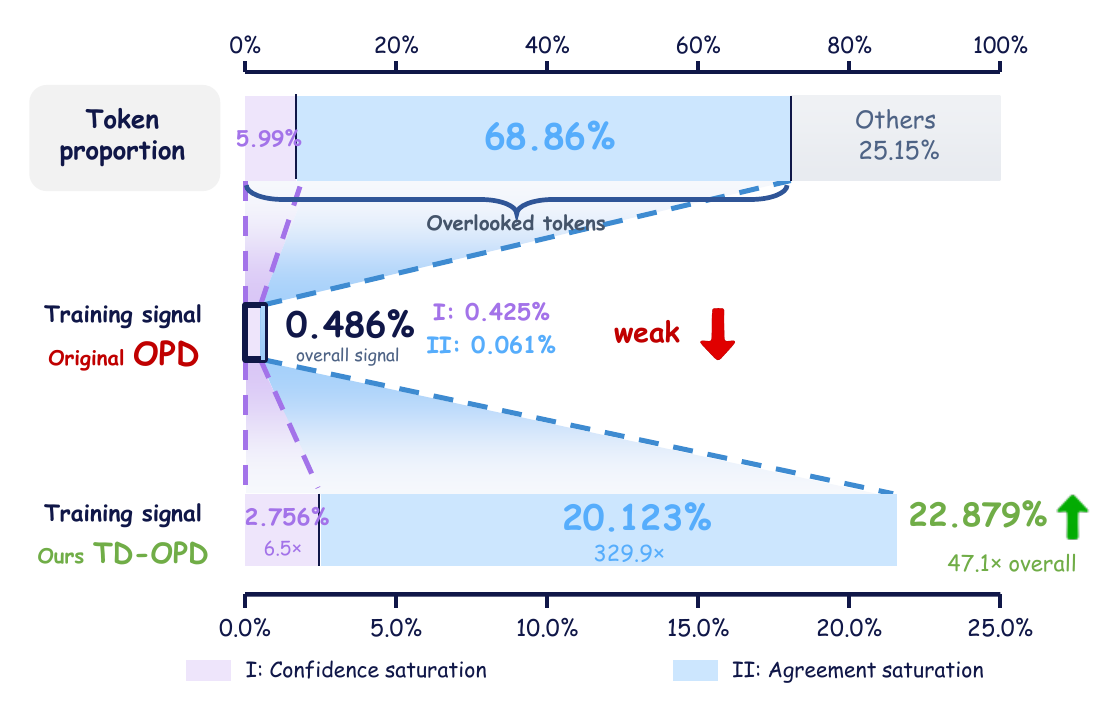}
    \caption{Weak signals in OPD.}
    \label{fig:weak_signals}
\end{wrapfigure}

\textbf{Prevalence and Prefix Sensitivity.}
We use two disjoint full-view regions as diagnostic pools:
$
R_{\mathrm{I}}=\{q_t \ge 0.95,\; A_t^H < -0.05\},
R_{\mathrm{II}}=\{|A_t^H| \le 0.05\},
$
where $R_{\mathrm{I}}$ captures highly confident predictions that the teacher disfavors, and $R_{\mathrm{II}}$ captures near-agreement predictions. The second region is intentionally broad: it contains both routine local continuations and prefix-dependent predictions whose full-history advantages are similarly small. Across early, middle, and late checkpoints of standard OPD, $R_{\mathrm{I}} \cup R_{\mathrm{II}}$ contains 74.85\%, 73.32\%, and 73.24\% of response tokens, while contributing only 0.486\%, 0.476\%, and 0.479\% of the full-view own-logit signal mass. These statistics identify a large pool of weakly supervised positions whose learning value is not resolved by the full-view signal alone.

We next use a controlled prefix intervention to probe which of these weak signals depend on access to the reasoning history. Keeping the sampled targets and full-view region assignments fixed, we alter the student's access to earlier reasoning and remeasure the resulting token-level signals. The signal share of the same $R_{\mathrm{I}} \cup R_{\mathrm{II}}$ tokens rises to 22.88\%, 16.39\%, and 17.70\% across the three checkpoints. Within $R_{\mathrm{II}}$, 48.85\%, 46.70\%, and 45.58\% of tokens show a positive signal gain. This response reveals a substantial prefix-sensitive subset within the broad near-agreement region: positions that appear indistinguishable from routine continuations under the complete history can acquire substantially stronger training signals when access to earlier reasoning is perturbed. Figure~\ref{fig:weak_signals} illustrates the early checkpoint; Appendix~\ref{app:signal_analysis} gives the intervention details and results. 


\FloatBarrier

%% file: Section/03-TD-OPD.tex
\section{Trajectory Dropout for OPD}
\subsection{Full-Context Rollouts with Masked-View Training}
\label{sec:td}

The controlled intervention above shows that weak full-view signals can conceal substantial sensitivity to the available reasoning history. This suggests treating the student's access to its own trajectory as a controllable training-time variable: selectively reducing prefix support can expose learning signals that are weak under the complete history, while the original on-policy trajectory and the teacher's full-prefix targets remain unchanged. We instantiate this idea as \emph{Trajectory Dropout} (TD). 
Applied to standard OPD, we call the resulting method TD-OPD; we use OPD+TD in comparative settings to emphasize its plug-in nature.

\begin{figure}[!htbp]
 \centering
 \includegraphics[width=0.80\linewidth]{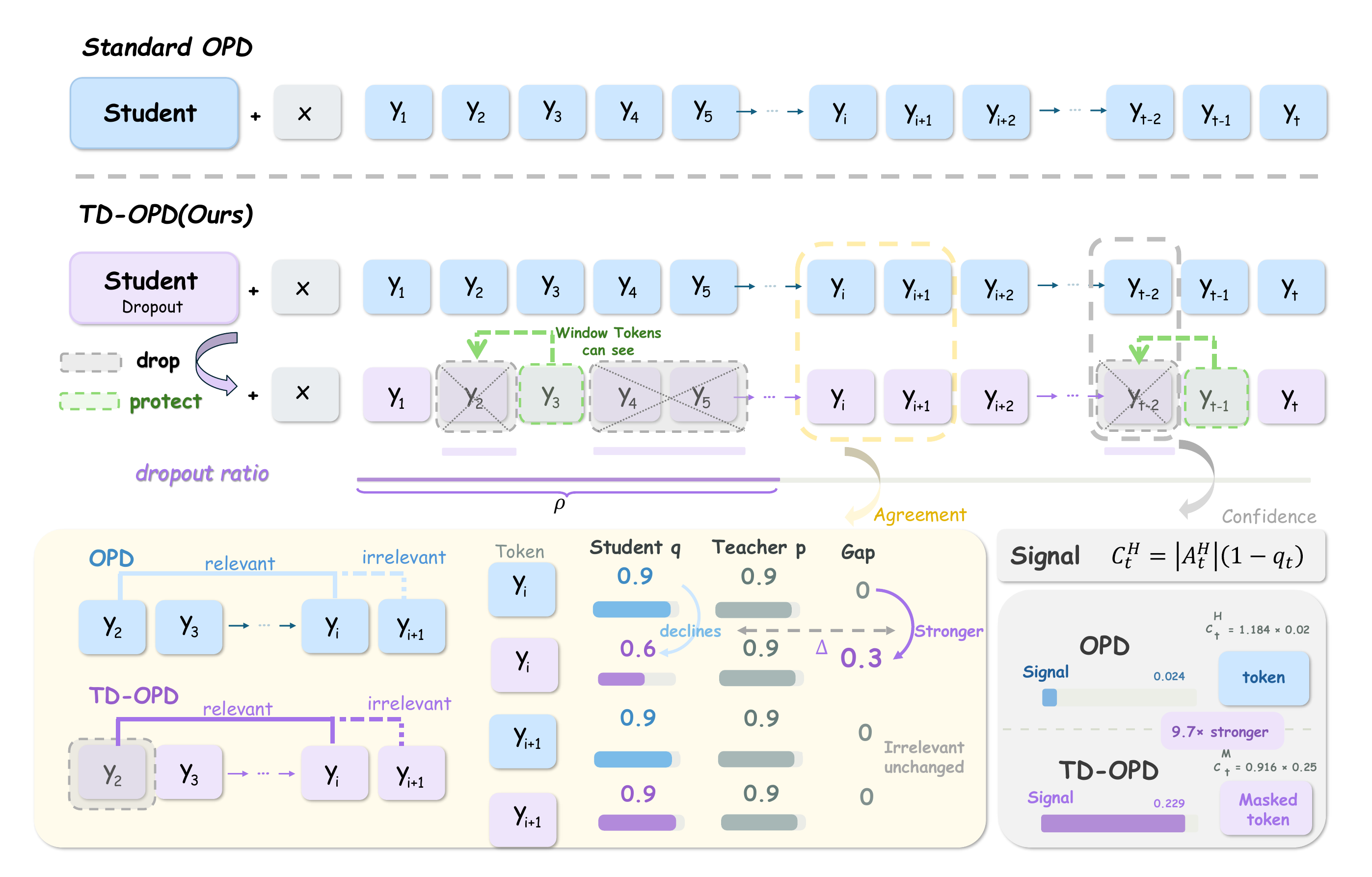}
 \caption{TD-OPD training.
 The student first completes a standard full-context rollout.
 During training, it reprocesses the saved trajectory with restricted attention to selected spans.
 The teacher scores every target on its full causal prefix.
 Old and current student scores use the same saved mask.
 These are fixed and trainable parameter states of the same student.
 All response targets receive supervision; evaluation uses full context.}
 \label{fig:td-overview}
\end{figure}

TD separates standard full-context trajectory generation from dropout-based training. The student first samples a complete trajectory
$y \sim Q_{\bar{\theta}}(\cdot \mid x)$
using standard autoregressive generation. Each token is generated from the prompt and the full preceding response, with standard causal attention throughout. The completed trajectory is then saved for training. During the subsequent student training passes, TD modifies access to selected parts of the saved reasoning history, while the teacher continues to score the same target tokens on their complete causal prefixes
(Figure~\ref{fig:td-overview}).

\textbf{Training-time Dropout.}
During training, we sample contiguous spans from the completed response.
Their total length covers a fraction $\rho$ of response source tokens.
These spans define an attention mask $M$, which is kept fixed for the update.
The student reprocesses the saved trajectory under $M$ to compute its training scores, and the teacher scores the same target tokens on their full causal prefixes $h_t$.
The prompt remains visible, every response target contributes to the loss.

For a selected span $[a,b)$, the mask blocks direct attention to its source positions
from targets $t\geq b$.
Predictions within the span keep causal access to its earlier tokens.
An optional protection window of length $w$ moves the boundary to $t\geq b+w$.
The next $w$ targets can then still attend to the span, preserving nearby prediction cues.
The mask applies in every attention head and layer.
The ratio $\rho$ measures selected source positions; the number of blocked edges also depends on their locations.
The experimental setup and Appendix~\ref{app:mask} give the hyperparameter settings and sampling details.

\FloatBarrier
\subsection{Cross-View Learning Signals}
\label{sec:mechanism}

\textbf{Cross-view Training Objective.}
TD changes the student's training view while keeping both the sampled trajectory and the teacher targets fixed in the full-context view. Let $r_{\theta,t}=R_\theta(y_t\mid h_t;M)$
denote the student's probability of the sampled token under the training mask $M$. Here $Q_\theta$ and $R_\theta$ share the same parameters and differ only in their attention access. Both the current student and its pre-update snapshot score the saved trajectory under the same mask. We therefore replace the student-side scores in the OPD surrogate with their masked-view counterparts: $a_t^M(\theta) = \operatorname{sg}\!\left[\log p_t - \log r_{\theta,t}\right], w_t^M(\theta) = \frac{r_{\theta,t}}{r_{\bar{\theta},t}},
\label{eq:masked-advantage}$ giving
\begin{equation}
L_{\mathrm{TD}}(\theta) = -\frac{1}{N}\sum_t w_t^M(\theta) a_t^M(\theta).
\label{eq:td-loss}
\end{equation}

The numerator and denominator of $w_t^M$ use the same mask, so the ratio tracks parameter changes within the student's training view. This does not alter how the trajectory is sampled: the response still comes from the full-context student, and the teacher score $p_t$ is still computed on the complete causal prefix. Thus, TD changes the view under which the student is optimized while leaving the rollout and supervision targets unchanged. At the start of an update, the ratio equals one; Appendix~\ref{app:local_regression} gives the corresponding local regression interpretation. We next analyze how this change in the student view modifies the token-level update.

Compare both views before the update, at $\theta=\bar\theta$.
Write $r_t=r_{\bar\theta,t}$ and $\delta_t=\log q_t-\log r_t$.
A positive $\delta_t$ means that the selected attention mask lowers the sampled token's probability.
We use $\delta_t$ as a measure of sensitivity to this mask.
The masked advantage satisfies
\begin{equation}
 A_t^M=\log p_t-\log r_t=A_t^H+\delta_t.
 \label{eq:adv-decomposition}
\end{equation}
Here $A_t^M$ is the raw value of $a_t^M(\theta)$ at the update start.
Dropout also changes the score vector to
$s_t^M=\left.\nabla_\theta\log r_{\theta,t}\right|_{\bar\theta}$.
For $u_t^M=A_t^Ms_t^M$, the update difference is
\begin{equation}
 u_t^M-u_t^H=
 \underbrace{A_t^H(s_t^M-s_t^H)}_{\text{changed student sensitivity}}
 +\underbrace{\delta_t s_t^M}_{\text{added advantage}}.
 \label{eq:update-decomposition}
\end{equation}
TD thus changes both factors that shape the learning signal.

\textbf{Making Existing Corrections Stronger.}
For a highly confident token, a lower $r_t$ increases the own-logit derivative $1-r_t$.
This can strengthen a downward correction even as the negative advantage shrinks.
For fixed $p\in(0,1)$, define $F_p(v)=(1-v)\log(v/p)$ on $v\in(p,1)$.
The curve has a unique maximum at $v_p^\star$ and decreases after it.
Thus, with $C_t^M=(1-r_t)\log(r_t/p_t)$,
\begin{equation}
 v_{p_t}^\star\leq r_t<q_t<1
 \quad\Longrightarrow\quad
 C_t^M>C_t^H.
 \label{eq:confidence-gain}
\end{equation}
This condition is sufficient rather than necessary; strong masking can
occasionally move $r_t$ below $p_t$ and reverse the correction direction.
We quantify these rare cases and a simple sign-preserving fallback in
Appendix~\ref{app:sign-reversal}.

\textbf{Disambiguating Near-agreement Predictions.}
Under exact full-view agreement, $A_t^H=0$, and Equation~\ref{eq:adv-decomposition} gives $A_t^M=\delta_t$, so the masked-view update becomes $u_t^M=\delta_t s_t^M$. Predictions that are indistinguishable under the full-view advantage can therefore separate under a prefix intervention according to their sensitivity $\delta_t$. Positions with $\delta_t\approx0$ remain weakly updated, whereas prefix-sensitive positions acquire an additional update through the induced advantage. This provides a training signal for differentiating positions within the broad near-agreement region that full-view agreement alone cannot resolve.

%% file: Section/03-Experiment.tex
\section{Experiments}
\label{sec:experiments}

\subsection{Models and Datasets}
\label{sec:experimental_setup}

\textbf{Models and Training Data.}
Our main setting uses Qwen3-4B-Instruct-2507 as the teacher and Qwen3-1.7B~\citep{yang2025qwen3}  as the student.
We also evaluate Qwen3-0.6B~\citep{yang2025qwen3} with the same teacher.
Both students use non-thinking mode.
\begin{wrapfigure}[12]{r}{0.4\textwidth}
    \centering
    \includegraphics[width=\linewidth]{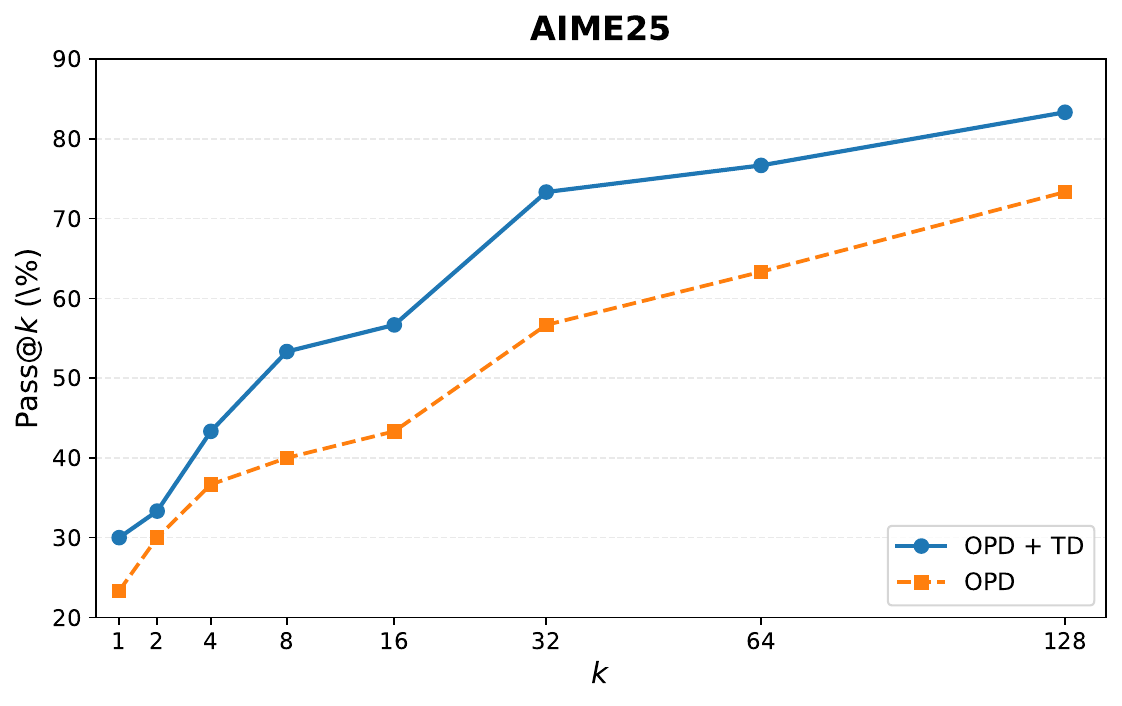}
    \caption{Pass@k performance of OPD and OPD+TD on AIME25}
    \label{fig:passk}
\end{wrapfigure}
We train on the teacher-correct English subset of DAPO-Math-17K~\citep{yu2025dapo}.
Each base method and its TD variant share the teacher, student initialization, training data, and chat template.

\textbf{Benchmarks.}
We evaluate AIME24, AIME25, AIME26, HMMT Feb26, and HMMT Nov25
using Avg@8, and OlympiadBench~\citep{he2024olympiadbench} using Avg@4. We additionally evaluate the code benchmark MBPP+~\citep{mbpp} and the science benchmark GPQA Diamond~\citep{gbqa} using Avg@4. All evaluations use standard full-prefix inference.
Dataset versions, sampling settings, and answer grading are given in
Appendix~\ref{app:exp-eval}.

\textbf{Implementation Details.}
The default dropout ratio is $\rho=0.2$.
For each trajectory, we randomly select one to eight response spans
whose total length matches the drop budget.
TD changes the student's prefix visibility during updates.
Rollout generation and teacher scoring use the complete causal prefix.
Causal access within each selected span is preserved.
All experiments use eight NVIDIA RTX 4090 GPUs.
Detailed hyperparameters are provided in Appendix~\ref{app:exp-config}.

\input{Section/Experiment_Table/main_table}

\subsection{Results}
\label{sec:main_results}

Table~\ref{tab:exp-main} reports the main Qwen3-1.7B results.
TD improves OPD from 28.27 to 32.78 in Avg, a gain of 4.5 percentage
points and a relative improvement of 15.9\%.
It also improves Fast-OPD from 29.48 to 33.76 and TRD from 18.76 to 22.68.
These correspond to relative gains of 14.6\% and 20.9\%, respectively.
All three methods improve on each of the six benchmarks when combined
with TD. We further evaluate TD under the thinking-on setting.
With thinking enabled, OPD achieves an Avg of 45.40, while OPD+TD
further improves it to 47.29, corresponding to a gain of 1.89 percentage
points and a relative improvement of 4.1\%. Figure~\ref{fig:passk} shows that OPD+TD consistently outperforms OPD on AIME25 in terms of Pass@k.  OPD+TD  also shows consistent improvements on the out-of-domain code benchmark MBPP+ and the science benchmark GPQA. The training dynamics and evaluation results for the 0.6B student are provided in Appendix~\ref{app:oter_results}. 



\subsection{Dropout Analysis}
\label{sec:dropout_analysis}

We use the main teacher--student pair and OPD for these ablations.
Each setting is evaluated on the six benchmarks with full-prefix inference.
The comparisons use a common training budget.



\begin{figure}[!htbp]

 \centering
 \includegraphics[width=0.8\linewidth]{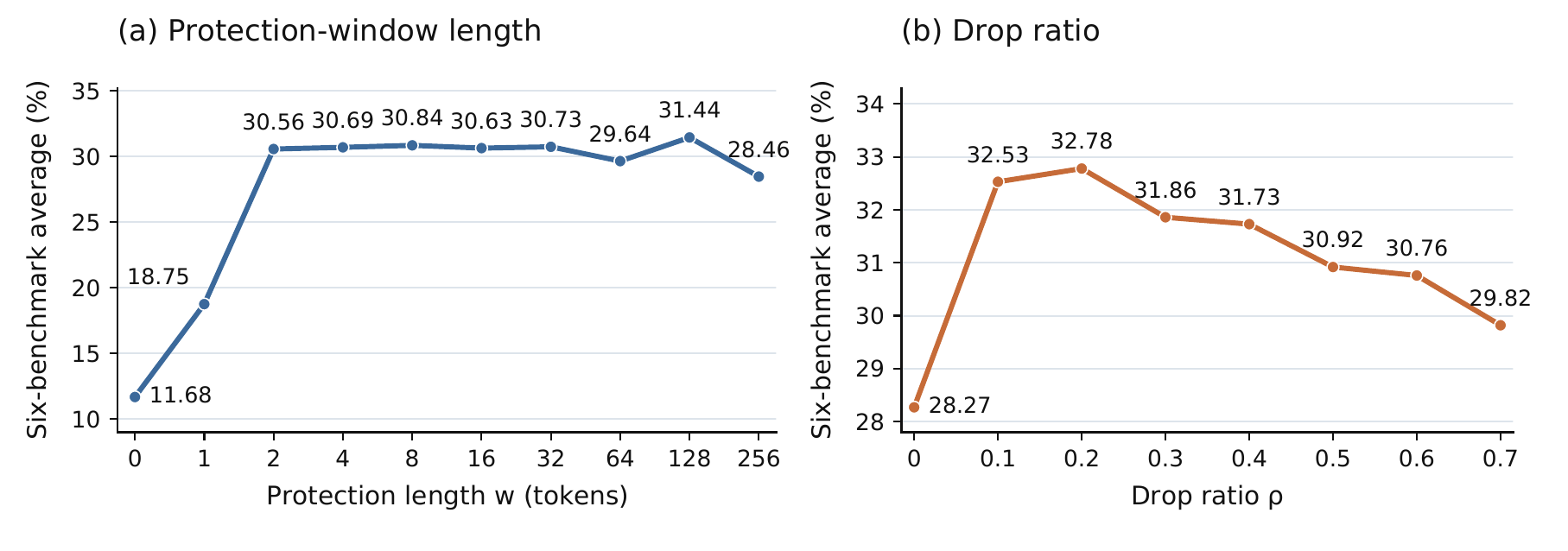}
 \caption{Sensitivity of TD-OPD to protection-window length and dropout ratio. Average performance across six mathematical reasoning benchmarks with varying (a) protection-window length \(w\) and (b) dropout ratio \(\rho\).}
 \label{fig:pro-dropout-ratio}
\end{figure}
\textbf{Dropout Ratio}. To investigate the impact of dropout ratio on performance, we report evaluation results trained under different dropout ratios. Figure~\ref{fig:pro-dropout-ratio}(b) shows that TD-OPD is robust to the dropout ratio under the current settings. The performance improves when the dropout ratio increases from 0 to 0.2 and then gradually decreases, with the best result achieved at 0.2. This suggests that a moderate dropout ratio is beneficial, whereas excessive dropout may lead to performance degradation.

\input{Section/Experiment_Table/teacher_ablation}
\textbf{Mask Granularity and Local Protection}. To study mask granularity, we replace contiguous spans with single-token
units while keeping the dropout ratio at $\rho=0.2$.
Without additional protection, accuracy drops sharply and training
becomes unstable.
This change also alters access near mask boundaries.
Span-wise TD preserves causal access within each selected span.
With single-token units, each selected token is masked from direct
attention starting at the next prediction position.
We propose that these repeated losses of local access contribute
to the degradation.

To test this explanation, we keep single-token masking at $\rho=0.2$
and add a protection window of length $w$ after each selected token.
The next $w$ prediction positions retain direct causal attention to
that token.
The token remains hidden from direct attention beyond this window.
The single-token sampling rule and all response targets are unchanged.
As in span-wise TD, masking is applied during training on completed
full-context rollouts.
We test $w\in\{1,2,4,8,16,32,64,128,256\}$, with $w=0$ as the reference.

Figure~\ref{fig:pro-dropout-ratio}(a) shows that even a short protection window leads to a substantial
recovery in accuracy.
The masking units remain single tokens throughout this comparison. This points to disrupted local access as an important source of the
degradation. The result supports preserving nearby prediction cues while restricting
direct access to earlier reasoning history.

\textbf{Role of Teacher Supervision}. To test whether student-side trajectory dropout alone can improve reasoning, we replace the stronger teacher with the Qwen3-1.7B student itself. The teacher uses full-context predictions, while TD is applied only to the student during training. As shown in Table~\ref{tab:teacher_ablation}, this self-distillation setting yields no improvement. It supports a complementary role for teacher supervision and
trajectory dropout: the stronger teacher supplies guidance, while
TD changes how that guidance is expressed in student updates. 




\subsection{Ablation and Analysis}
\textbf{Update Norm Controlled Analysis}. To rule out the possibility that TD's gain merely reflects a larger update magnitude, we conduct a gradient-norm-controlled paired one-step analysis and find that, with update norms strictly matched to the OPD baseline, the TD update still yields a significantly larger held-out teacher-alignment KL improvement than the plain OPD update ($+0.027$, 95\% CI $[0.022, 0.034]$, $p<0.001$), demonstrating that the advantage of TD is directional. It stems from a better learning signal rather than a larger step size; see Appendix~\ref{app:norm-controlled} for details.

\textbf{Channel Decomposition}. A further ablation of the same experiment shows that neither change alone
suffices: at matched update norms, hiding the removed spans only in the
student's forward computation ($-0.004$), or computing the per-token
training signal only under the masked view ($-0.002$), each performs worse
than the full-context OPD update, and only the two changes applied
together---the actual TD-OPD update---bring the improvement ($+0.027$); the
benefit therefore requires that what the student sees and how its training
signal is computed are masked \emph{consistently}, rather than either one
alone (Appendix~\ref{app:norm-controlled}).

%% file: Section/Experiment_Table/main_table.tex

\begin{table*}[t]
\centering
\caption{
Performance (\%) of Qwen3-1.7B under non-thinking and thinking inference.
Panel (b) enables thinking only at inference.
Math Avg is the unweighted mean over the six mathematical benchmarks,
excluding MBPP+ and GPQA Diamond.
TD variants are shaded, with absolute gains (percentage points) over their
counterparts without TD shown below each score.
Bold and underlined values mark the best and second-best reported scores
within each panel and column.
}
\label{tab:exp-main}

\begingroup
\small
\setlength{\tabcolsep}{1.6pt}
\renewcommand{\arraystretch}{1.15}
\renewcommand{\tabularxcolumn}[1]{m{#1}}

\newcommand{\TDcell}[2]{%
  \shortstack[c]{#1\\[-1pt]
  {\scriptsize\textcolor{blue!65!black}{(#2)}}}%
}

\newcommand{\OODmissing}{%
  \textcolor{black!40}{\texttt{--}}%
}

\begin{tabularx}{\linewidth}{
  l
  *{9}{>{\centering\arraybackslash}X}
}
\toprule

& \multicolumn{7}{c}{\textbf{Mathematical reasoning}}
& \multicolumn{2}{c}{
  \shortstack{\textbf{Out-of-domain}\\\textbf{generalization}}
} \\

\cmidrule(lr){2-8}
\cmidrule(l){9-10}

Method
& AIME24
& AIME25
& AIME26
& Olymp.
& \shortstack{HMMT\\Feb26}
& \shortstack{HMMT\\Nov25}
& \shortstack{Math\\Avg}
& MBPP+
& \shortstack{GPQA\\-D}
\\

\midrule

\multicolumn{10}{@{}l@{}}{
  \textbf{(a) Non-thinking inference}
} \\

\addlinespace[3pt]

Student
& 12.50
& 9.58
& 7.50
& 38.56
& 6.44
& 4.58
& 13.19
& 55.62
& 27.90
\\

SFT
& 23.33
& 19.58
& 16.25
& 46.63
& 12.50
& 6.67
& 20.83
& 56.08
& 28.91
\\

KD
& 23.75
& 21.25
& 15.42
& 48.07
& 12.88
& 7.50
& 21.48
& 56.48
& 28.66
\\

GRPO
& 24.58
& 22.08
& 15.83
& 48.74
& 14.39
& 9.58
& 22.54
& 57.21
& 29.55
\\

\midrule

OPD
& 34.58
& 22.92
& 23.33
& 55.41
& 20.08
& 13.33
& 28.27
& \underline{60.19}
& 31.57
\\

\rowcolor{blue!7}
OPD + TD
& \TDcell{\textbf{40.42}}{+5.83}
& \TDcell{\textbf{30.42}}{+7.50}
& \TDcell{\underline{25.83}}{+2.50}
& \TDcell{\underline{58.07}}{+2.67}
& \TDcell{\underline{22.35}}{+2.27}
& \TDcell{\underline{19.58}}{+6.25}
& \TDcell{\underline{32.78}}{\textbf{+4.50}}
& \TDcell{\textbf{60.32}}{+0.13}
& \TDcell{\textbf{35.73}}{+4.16}
\\

\addlinespace[3pt]

TRD
& 19.17
& 19.58
& 12.92
& 44.19
& 12.12
& 4.58
& 18.76
& 54.70
& 27.78
\\

\rowcolor{blue!7}
TRD + TD
& \TDcell{26.25}{+7.08}
& \TDcell{22.08}{+2.50}
& \TDcell{17.50}{+4.58}
& \TDcell{48.67}{+4.48}
& \TDcell{13.64}{+1.52}
& \TDcell{7.92}{+3.33}
& \TDcell{22.68}{\textbf{+3.92}}
& \TDcell{55.03}{+0.33}
& \TDcell{29.92}{+2.14}
\\

\addlinespace[3pt]

Fast-OPD
& 36.67
& \underline{27.08}
& 22.50
& 56.74
& 19.70
& 14.17
& 29.48
& 59.26
& 33.08
\\

\rowcolor{blue!7}
Fast-OPD + TD
& \TDcell{\underline{40.00}}{+3.33}
& \TDcell{\textbf{30.42}}{+3.33}
& \TDcell{\textbf{29.17}}{+6.67}
& \TDcell{\textbf{59.48}}{+2.74}
& \TDcell{\textbf{23.11}}{+3.41}
& \TDcell{\textbf{20.42}}{+6.25}
& \TDcell{\textbf{33.76}}{\textbf{+4.29}}
& \TDcell{60.05}{+0.79}
& \TDcell{\underline{34.47}}{+1.39}
\\

\addlinespace[6pt]
\midrule

\multicolumn{10}{@{}l@{}}{
  \textbf{(b) Thinking inference}\quad
  {\footnotesize\textcolor{black!70}{
    OPD variants trained in non-thinking mode
  }}
} \\

\addlinespace[3pt]

Student
& 53.33
& 34.17
& \underline{38.75}
& 66.81
& 28.41
& 27.92
& 41.57
& 66.93
& 36.74
\\

OPD
& \underline{55.42}
& \underline{41.67}
& \textbf{43.33}
& \underline{68.37}
& \underline{30.30}
& \underline{33.33}
& \underline{45.40}
& \underline{68.39}
& \underline{39.14}
\\

\rowcolor{blue!7}
OPD + TD
& \TDcell{\textbf{57.08}}{+1.66}
& \TDcell{\textbf{42.50}}{+0.83}
& \TDcell{\textbf{43.33}}{+0.00}
& \TDcell{\textbf{69.63}}{+1.26}
& \TDcell{\textbf{34.09}}{+3.79}
& \TDcell{\textbf{37.08}}{+3.75}
& \TDcell{\textbf{47.29}}{\textbf{+1.89}}
& \TDcell{\textbf{69.58}}{+1.19}
& \TDcell{\textbf{41.54}}{+2.40}
\\

\bottomrule
\end{tabularx}

\endgroup
\end{table*}

%% file: Section/Experiment_Table/teacher_ablation.tex
\begin{wraptable}{r}{0.42\textwidth}
    \centering
    \small
    \setlength{\abovecaptionskip}{0pt}
    \setlength{\belowcaptionskip}{5pt}
    \setlength{\tabcolsep}{3pt}
    \renewcommand{\arraystretch}{1.12}

    \caption{
        Teacher ablation under OPD+TD. The student is Qwen3-1.7B throughout.
        }
    \label{tab:teacher_ablation}

    \begin{tabular*}{\linewidth}{@{\extracolsep{\fill}}lccc@{}}
        \toprule
        Teacher & AIME24 & AIME25 & AIME26 \\
        \midrule
        None (initial)
            & 12.50 & 9.58 & 7.50 \\
        Qwen3-1.7B
            & 10.83 & 10.83 & 7.50 \\
        \addlinespace[2pt]
        \begin{tabular}[c]{@{}l@{}}
            Qwen3-4B-\\Instruct-2507
        \end{tabular}
            & \textbf{40.42}
            & \textbf{30.42}
            & \textbf{25.83} \\
        \bottomrule
    \end{tabular*}
\end{wraptable}

%% file: Section/Related_work.tex
\section{Related Work}

\paragraph{On-policy distillation.}
On-policy distillation (OPD) trains students on self-generated trajectories using dense teacher feedback \citep{gu2024minillm,agarwal2024gkd}. Recent work improves its efficiency or supervision quality through rollout scheduling \citep{fastopd2026,adaptivefastopd2026}, token-level selection or reweighting \citep{tip2026,fireopd2026,taopd2026,prefixteach2026}, and trajectory-level refinement or guidance \citep{renio2026,trd2026,topd2026}. Other methods create asymmetric supervision by giving the teacher privileged or alternative context unavailable to the student \citep{opsd2026,opcd2026,lin2026ccopd}. These approaches primarily modify the objective, weighting, trajectory, or teacher context. Our work instead studies how the student's own reasoning prefix shapes the resulting update: TD keeps the rollout and full-prefix teacher targets fixed, while modifying only the student's access to its reasoning history during optimization.

\paragraph{Context masking and asymmetric distillation.}
Training-time context masking has been explored for data augmentation and reasoning distillation \citep{gao2025tokendropout,chen2026pretrainaug,reflectioncoder2025,trsd2026,maskeddistillation2026}. Most closely related, Hide to See \citep{hidetosee2026} masks reasoning prefixes to reduce textual shortcuts in multimodal distillation, while visual on-policy methods pair clean teacher views with corrupted student inputs \citep{s2vopd2026,nopd2026}. TD differs in \emph{when and why} this asymmetry is introduced: the student first generates a standard full-context on-policy trajectory, and masking is applied only when replaying that trajectory for optimization. Thus, TD preserves both the rollout distribution and full-prefix teacher supervision, using the masked student view to strengthen existing corrections or introduce additional signals at positions that receive weak updates under shared-prefix OPD. 

%% file: Section/04-Conclustion.tex
\section{Conclusion}
\label{sec:conclusion}

We studied how dense teacher supervision under a shared reasoning prefix can produce highly uneven update strengths in OPD.
We identified \emph{prefix-induced supervision attenuation} (PISA), which weakens learning signals through confidence saturation and agreement saturation.
We proposed \emph{trajectory dropout} (TD), which randomly masks reasoning history during student updates.
TD strengthens corrective responses and selectively introduces recovery signals for prefix-dependent predictions.
Experiments across teacher--student pairs and six reasoning benchmarks show substantial gains over standard OPD.
TD also improves existing variants, including TRD and Fast-OPD.
Ablations examine the effect of the dropout ratio and show that preserving local prefix continuity supports stable training.

%% file: Appendix/other_results.tex
\section{Training Dynamics and Results}
\label{app:oter_results}
\FloatBarrier

\begin{figure}[!htbp]

 \centering
 \includegraphics[width=0.8\linewidth]{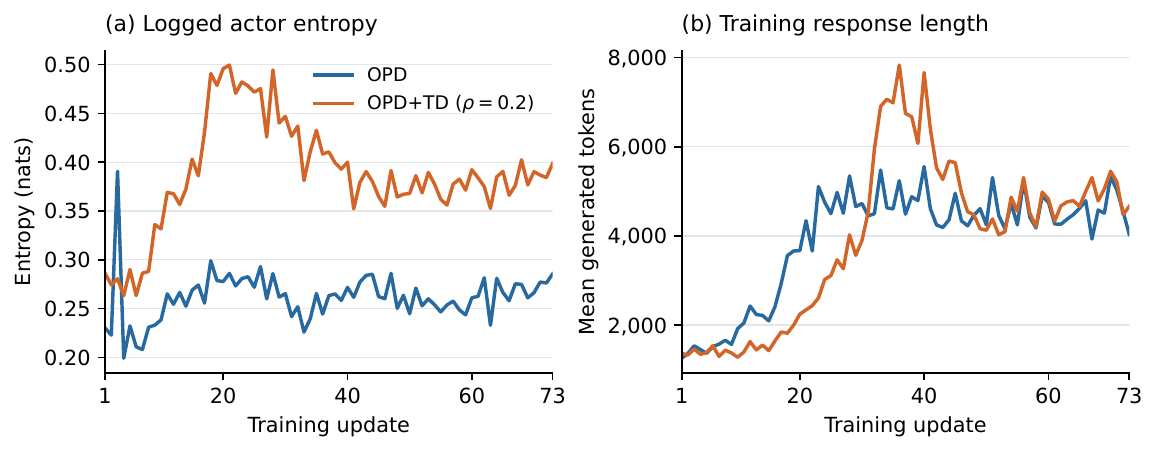}
 \caption{Training dynamics of Qwen3-1.7B with OPD and OPD+TD ($\rho$= 0.2).}
 \label{fig:training-dynamics}
\end{figure}

\textbf{Training dynamics.}
Figure~\ref{fig:training-dynamics} (a) shows the student entropy recorded
during training.
The TD-OPD curve shows higher student entropy in these traces.
This indicates less concentrated token distributions and is consistent
with TD reducing confidence on predictions affected by prefix masking.
Mean training response lengths are close, at 3,956 and 3,933 tokens. Figure~\ref{fig:training-dynamics} (b) shows OPD+TD  a larger length peak near the middle of training.

\textbf{Qwen3-0.6B results.} Table~\ref{tab:exp-main-06b} shows that with the Qwen3-0.6B student model, adding the TD method still demonstrates a trend of improving accuracy across the six datasets.

\textbf{Robustness Across Training Seeds}. To examine sensitivity to training randomness, we train OPD+TD with
three independent seeds. The six-benchmark averages are 32.78, 32.91,
and 32.46, yielding $32.72 \pm 0.23$ (mean $\pm$ standard deviation).
The small run-to-run variation suggests that the performance of TD is
stable across training seeds. All three runs remain substantially above
the OPD result (28.27) reported in the main experiment.

\input{Section/Experiment_Table/main_0.6B}


%% file: Section/Experiment_Table/main_0.6B.tex
\begin{table}[t]
\centering
\caption{
Mathematical reasoning performance of Qwen3-0.6B
in non-thinking mode.
All scores are percentages.
Avg is the unweighted mean over the six benchmarks.
TD variants are shaded.
The best scores are bold, and the second-best scores are underlined.
}
\label{tab:exp-main-06b}

\begingroup
\small
\setlength{\tabcolsep}{3pt}
\renewcommand{\arraystretch}{1.18}

\begin{tabularx}{\linewidth}{
    @{}l*{7}{>{\centering\arraybackslash}X}@{}
}
\toprule
Method
& AIME24
& AIME25
& AIME26
& Olymp.
& \shortstack{HMMT\\Feb26}
& \shortstack{HMMT\\Nov25}
& Avg \\
\midrule

Student
& 1.67
& 2.50
& 0.83
& 16.37
& 0.76
& 3.75
& 4.31 \\

SFT
& 5.00
& 7.50
& 4.17
& 26.74
& 1.89
& 2.92
& 8.04 \\

KD
& 4.17
& 7.08
& 5.00
& 27.15
& 3.03
& 2.92
& 8.22 \\

GRPO
& 8.33
& 15.42
& 10.00
& 35.41
& 7.95
& 6.25
& 13.89 \\

\midrule

OPD
& \textbf{13.33}
& 16.67
& 12.08
& 41.33
& 7.58
& 5.42
& 16.07 \\

\rowcolor{blue!7}
OPD + TD
& \underline{12.92}
& \underline{17.92}
& \textbf{15.00}
& 42.41
& \underline{11.74}
& \textbf{8.75}
& \underline{18.12}\,
 \\

TRD
& 4.58
& 7.08
& 4.17
& 26.93
& 3.79
& 2.92
& 8.24 \\

\rowcolor{blue!7}
TRD + TD
& 5.42
& 8.75
& 4.58
& 27.11
& 5.30
& 4.58
& 9.29\,
  \\

Fast-OPD
& 8.75
& 17.50
& \underline{12.50}
& \underline{45.52}
& 10.23
& 6.25
& 16.79 \\

\rowcolor{blue!7}
Fast-OPD + TD
& 11.67
& \textbf{20.42}
& 11.67
& \textbf{45.63}
& \textbf{13.64}
& \underline{8.33}
& \textbf{18.56}\,
 \\

\bottomrule
\end{tabularx}

\endgroup
\end{table}

%% file: Appendix/experimental-configurations.tex
\section{Experimental Configurations}
\label{app:exp-config}

Table~\ref{tab:exp-config-opd-td} summarizes the core hyperparameters and data settings of the main TD-OPD experiments with the 1.7B student. To prevent potential data contamination, we explicitly removed overlapping examples between the training data and evaluation benchmarks during preprocessing. All other OPD experiments follow the same configuration unless otherwise stated in the main text.

\textbf{Overhead of TD.} TD is nearly free by construction: it introduces no additional rollouts, no
teacher-scoring passes, and no optimizer steps. So that, measured on back-to-back twin training runs with
identical configuration and the same attention backend, it slows training by
only $2.8\%$ wall-clock per step ($273.3\,\mathrm{s} \rightarrow
280.9\,\mathrm{s}$, $+7.6\,$s), an increase confined to the student update
and log-probability passes while the dominant rollout generation remains
entirely untouched.
\begin{table}[htbp]
\centering
\caption{Core experimental configurations for the 1.7B student (TD-OPD).}
\label{tab:exp-config-opd-td}
\begin{tabular}{lc}
\toprule
\textbf{Parameter} & \textbf{TD-OPD ($\rho=0.2$)} \\
\midrule
Student & Qwen3-1.7B \\
Teacher & Qwen3-4B-Instruct-2507 \\
Training data & DAPO, 9,755 teacher-correct English examples \\
Objective & K1 reverse-KL, policy gradient \\
Optimizer & AdamW \\
Learning rate / schedule & $10^{-6}$ / constant \\
Weight decay & 0.01 \\
Global batch / mini-batch size & 128 / 128 \\
Update epochs per rollout batch & 1 \\
Maximum prompt / response length & 2,048 / 16,384 tokens \\
Rollouts per prompt & 1 \\
Sampling temperature / top-$p$ & 1.0 / 1.0 \\
gradient-norm cap &  1.0 \\
Data seed & 42  \\
dropout ratio $\rho$ & 0.2 \\
TD span sampling & Random positions and lengths; up to 8 spans \\
TD application probability & 1.0 \\
Additional local protection & 0 tokens \\
\bottomrule
\end{tabular}
\end{table}

%% file: Appendix/evaluation_and_dataset.tex
\section{Evaluation Datasets and Protocol}
\label{app:exp-eval}

\paragraph{Dataset versions.}
We evaluate on six mathematical reasoning benchmarks and two
out-of-domain benchmarks, summarized in
Table~\ref{tab:eval-datasets}. AIME24, AIME25, and AIME26 each contain
the 30 problems from AIME~I and AIME~II in the corresponding year.
Olymp denotes the English, text-only competition-mathematics portion
of OlympiadBench, using the 675-question evaluation snapshot.
The AIME26 and HMMT sets use the corresponding MathArena releases.
For code generation, we evaluate on MBPP+, using the 378-task
EvalPlus version of the benchmark.
For scientific reasoning, we evaluate on GPQA Diamond, the
198-question Diamond subset of GPQA covering graduate-level questions
in biology, physics, and chemistry.

\begin{table}[t]
\centering
\small
\setlength{\tabcolsep}{4pt}
\renewcommand{\arraystretch}{1.12}
\caption{Evaluation datasets and samples per problem ($n$).
Question counts refer to the evaluated versions.}
\label{tab:eval-datasets}
\begin{tabularx}{\linewidth}{@{}lXrr@{}}
\toprule
Benchmark & Version / subset & Questions & $n$ \\
\midrule
AIME24 & AIME I and II, 2024 & 30 & 8 \\
AIME25 & AIME I and II, 2025 & 30 & 8 \\
AIME26 & \href{https://huggingface.co/datasets/MathArena/aime_2026}{MathArena, AIME 2026} & 30 & 8 \\
Olymp & OlympiadBench, English text-only mathematics & 675 & 4 \\
HMMT Feb26 & \href{https://huggingface.co/datasets/MathArena/hmmt_feb_2026}{MathArena, February 2026} & 33 & 8 \\
HMMT Nov25 & \href{https://huggingface.co/datasets/MathArena/hmmt_nov_2025}{MathArena, November 2025} & 30 & 8 \\
MBPP+ & EvalPlus MBPP+ & 378 & 4 \\
GPQA Diamond & GPQA Diamond & 198 & 4 \\
\bottomrule
\end{tabularx}
\par\smallskip
\end{table}

\paragraph{Prompting and sampling.}
We use zero-shot prompts with the model's native chat template.
For the mathematical reasoning benchmarks, we request step-by-step
reasoning with the final answer in
\texttt{\textbackslash boxed\{\}}.
For GPQA Diamond, each question is presented with its four answer
choices, and the model is requested to provide its reasoning followed
by a final answer choice.
For MBPP+, the model is given the programming problem specification
and is asked to generate a Python solution satisfying the required
function interface.
Responses are sampled at temperature $1.0$ and top-$p=1.0$, with a maximum
of 32,768 generated tokens and a total context limit of 34,817 tokens.
For each problem, sample $j$ uses seed $42+j$, where $j=0,\ldots,n-1$.
Generation stops at the end-of-sequence token or the response-length limit.
TD is disabled during evaluation, and generation uses the full context.

\paragraph{Answer grading.}
For the mathematical reasoning benchmarks, we grade final answers using
rule-based mathematical equivalence checks.
Answer extraction first attempts the last boxed expression, followed by
an explicit \texttt{The answer is:} pattern and the Math-Verify parser.
OpenMathInstruct utilities and Math-Verify check numerical and symbolic
equivalence to the reference answer.
For OlympiadBench, OBJudge provides an additional check with tolerance
$10^{-8}$.

For GPQA Diamond, we extract the model's final multiple-choice answer
and compare it with the reference answer choice by exact match.
Responses from which no valid answer choice can be extracted are scored
as incorrect.
For MBPP+, functional correctness is evaluated using the EvalPlus
execution-based evaluator on the MBPP+ test suite. A generated solution
is counted as correct only if it passes the evaluation tests; syntax
errors, runtime errors, timeouts, and other execution failures are
scored as incorrect.

Across all benchmarks, unparseable answers and grading exceptions are
scored as incorrect. Length-truncated responses are graded as generated
and remain in the evaluation denominator. Intermediate reasoning
receives no separate score.

\paragraph{Metrics.}
Let $c_i$ be the number of correct responses among $n$ samples for
problem $i$, and let $N$ be the number of problems. We report
\begin{equation}
\operatorname{Avg}@n
=\frac{100}{Nn}\sum_{i=1}^{N}c_i,
\qquad
\operatorname{Pass}@n
=\frac{100}{N}\sum_{i=1}^{N}\mathbb{1}[c_i>0].
\end{equation}
Avg@$n$ measures average accuracy over independently sampled responses
without majority voting; Pass@$n$ measures the fraction of problems
solved at least once.
The six-benchmark Avg reported for mathematical reasoning is the
unweighted mean of Avg@$n$ across the six mathematical benchmarks.
MBPP+ and GPQA Diamond are reported separately and are not included in
this six-benchmark average.

%% file: Appendix/signal_analysis.tex

\section{Signal Analysis}
\label{app:signal_analysis}

We probe how trajectory dropout changes the training signal at early, middle,
and late checkpoints of standard OPD with a Qwen3-1.7B student
(steps 10, 40, and 70). At each checkpoint, we compare full-context scoring
with 20\% span dropout on the same next training batch of 128 trajectories,
using two independently seeded masks per trajectory.
Following the own-logit analysis in Section 2.2, we measure
$H_t^{\mathrm{full}}=|A_t^H|(1-q_t)$ and
$H_t^{M}=|A_t^M|(1-r_t)$.
Their sums measure sampled-token logit signal mass before the update.
We identify confidence-saturated tokens by $q_t\geq0.95$ and $A_t^H<-0.05$,
and agreement-saturated tokens by $|A_t^H|\leq0.05$.
These disjoint groups are defined in the full view and held fixed across views.

Weak signals are prevalent at all three audited stages.
Together, the two groups contain 74.85\%, 73.32\%, and 73.24\% of response
tokens, yet account for only 0.486\%, 0.476\%, and 0.479\% of total full-view
signal mass, respectively.
Under dropout, the signal shares of these same tokens rise to
22.88\%, 16.39\%, and 17.70\%. Their absolute signal mass increases by 67.67, 45.36, and 49.78 times;
these factors also account for the change in total signal mass between views.
Confidence-saturated tokens contribute 8.08\%, 8.55\%, and 8.01\% of the
net increase in total signal mass, while agreement-saturated tokens
contribute 65.92\%, 58.03\%, and 59.53\%.

Positive signal gain occurs in a substantial fraction of the original weak tokens.
50.61\%, 48.33\%, and 47.29\% of these tokens
contribute positive gain at the early, middle, and late checkpoints.
Within the confidence group, the corresponding fractions are 70.85\%,
65.58\%, and 66.41\%; within the agreement group, they are 48.85\%,
46.70\%, and 45.58\%.
These observations are consistent with the proposed mechanism: changing
prefix access can strengthen the signals of predictions that provide
little supervision in the full view.
Among confidence-saturated tokens retaining a downward correction, signal
mass still increases by 2.80, 2.19, and 2.20 times.

%% file: Appendix/Norm-Controlled-Comparison.tex
\section{Norm-Controlled Comparison of Update Directions}
\label{app:norm-controlled}

In standard OPD, the student always fits the teacher distribution under the
full context. Because changing the student's visible context also changes
the magnitudes of the gradients and of the resulting parameter updates, the
gain from OPD+TD could in principle arise merely from a difference in update
magnitude (i.e., an effective learning rate) rather than from a better
learning signal.

To rule out this explanation, we perform a paired one-step update
comparison on the same student model: for each training batch,
one OPD update and one OPD+TD update are computed from the identical weights
under exactly the same optimizer configuration (AdamW, learning rate
$1\times10^{-6}$, global gradient-norm clipping at $1.0$), and each update's
parameter displacement is rescaled to the same norm as that batch's OPD
update (three radii: $0.25\times$, $0.5\times$, and $1.0\times$; the maximum
realized relative deviation of the rescaled norms is $0.12\%$) before being
applied. The experiment covers 16 training batches; the effect is evaluated
on 128 held-out GSM8K reference solutions with no overlap with any training
data, and the metric is the change in the full-vocabulary forward KL
divergence $D_{\mathrm{KL}}(p_{\mathrm{teacher}} \,\|\, p_{\mathrm{student}})$
at 16 pre-fixed positions per question.

\begin{figure}[htbp]
\centering
\includegraphics[width=\linewidth]{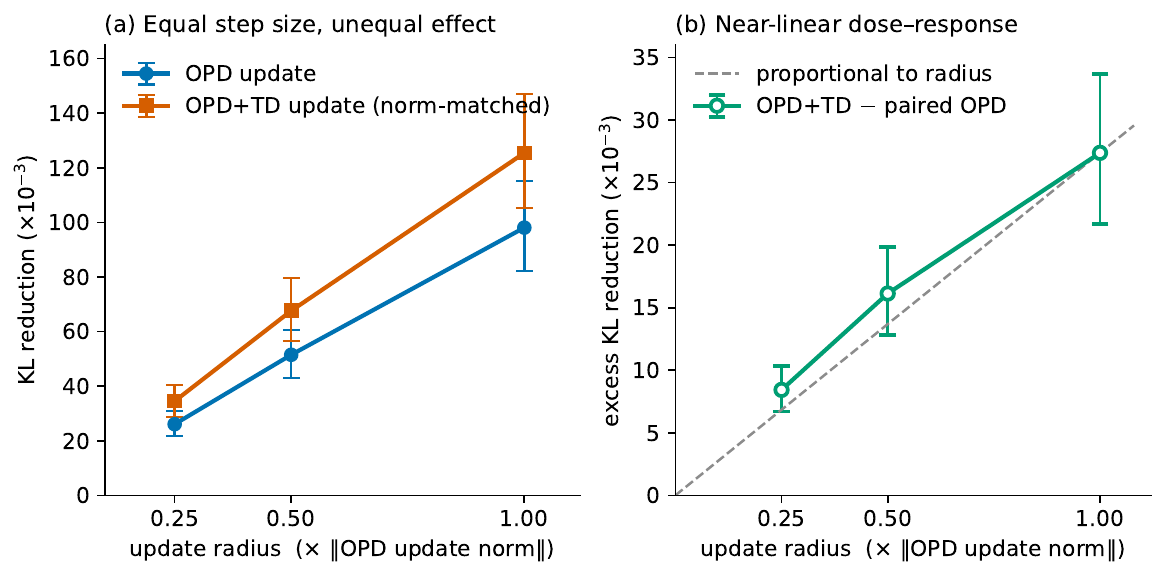}
\caption{Paired one-step updates under matched norms. (a) Absolute KL
improvement of the OPD update and the norm-matched OPD+TD update across the
three radii: equal step size, unequal effect---the OPD+TD update is higher
at every radius. (b) The excess improvement of OPD+TD over the paired OPD
update grows nearly linearly with the radius (dashed line: proportional
reference), indicating a directional rather than step-size effect. Error
bars are 95\% bootstrap CIs; the excess is positive in 16/16 training
batches at every radius.}
\label{fig:matched-norm}
\end{figure}

The results show that, under strictly aligned update norms, the OPD+TD
update still yields a significantly larger KL reduction than the paired OPD
update ($+0.027$, 95\% CI $[0.022, 0.034]$; with the training batch as the
statistical unit, 10{,}000 crossed batch$\times$question bootstrap resamples,
$p<0.001$ after multiple-comparison correction). The excess gain grows
nearly linearly with the update norm ($+0.008$, $+0.016$, and $+0.027$ at
the three radii), indicating that it stems from the alignment between the
update direction and the validation objective rather than from the distance
traveled (Figure~\ref{fig:matched-norm}, Table~\ref{tab:matched-norm}).

\begin{table}[htbp]
\centering
\caption{Excess KL improvement of the OPD+TD update over the paired OPD
update on held-out validation, at matched update norms (mean with 95\%
bootstrap CI). The top three rows use the three matched radii; the bottom
two rows are the unrescaled natural-norm update and the unpreconditioned
fixed-step-size gradient-update control.}
\label{tab:matched-norm}
\begin{tabular}{lc}
\toprule
Setting & Excess KL improvement \\
\midrule
Radius $0.25\times\|\Delta\theta_{\mathrm{OPD}}\|$ & $+0.008\;[0.007, 0.010]$ \\
Radius $0.5\times\|\Delta\theta_{\mathrm{OPD}}\|$  & $+0.016\;[0.013, 0.020]$ \\
Radius $1.0\times\|\Delta\theta_{\mathrm{OPD}}\|$  & $+0.027\;[0.022, 0.034]$ \\
\midrule
Unrescaled (natural norm) & $+0.027\;[0.021, 0.033]$ \\
Unpreconditioned fixed-step-size update & $+0.147\;[0.115, 0.182]$ \\
\bottomrule
\end{tabular}
\end{table}

Two additional observations rule out the remaining alternative explanations.
First, the natural (unrescaled) norm of the OPD+TD update is already close
to that of the OPD update (the result is nearly unchanged by the rescaling,
$+0.027$ versus $+0.027$), so OPD+TD does not rely on a larger step size.
Second, after replacing AdamW with an unpreconditioned fixed-step-size
gradient update, the directional advantage of OPD+TD remains significant
($+0.147$, $p<0.001$), so the conclusion does not depend on Adam-style
preconditioning.

\begin{figure}[htbp]
\centering
\includegraphics[width=\linewidth]{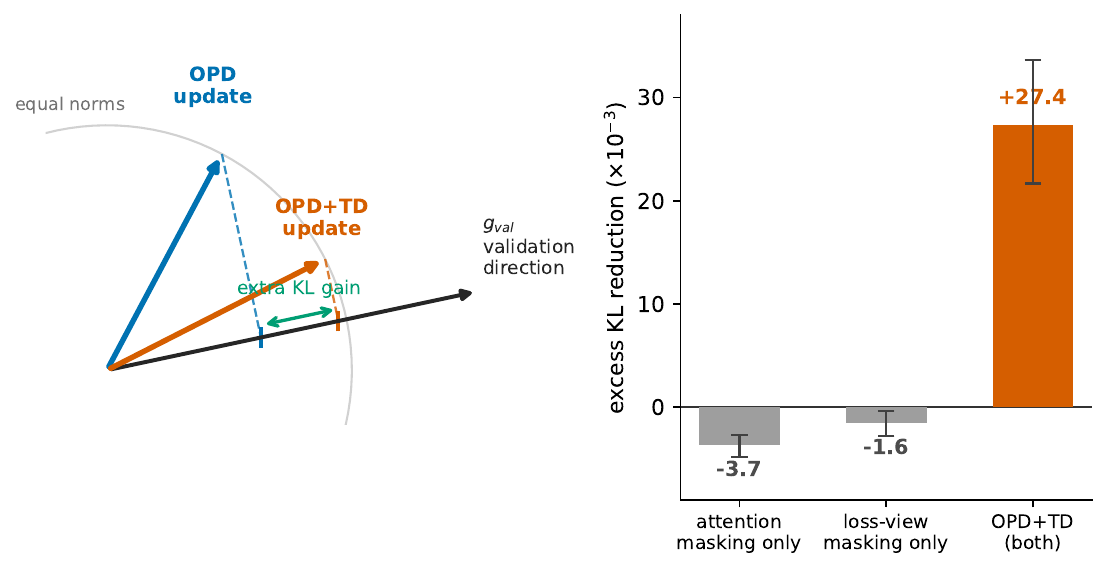}
\caption{Directional advantage: schematic and decomposition. (a) With equal
update norms, the difference between the OPD+TD and OPD updates lies in
their alignment with the validation-objective direction
$g_{\mathrm{val}}$ (a longer projection corresponds to a larger excess KL
improvement); angles are illustrative, not measured. (b) At matched norm
($1.0\times$), changing only a single channel (attention masking only,
$-0.004$; loss-view masking only, $-0.002$) performs worse than the
full-context OPD update, and only the full OPD+TD (both channels changed
jointly, $+0.027$) improves.}
\label{fig:direction-schematic}
\end{figure}

Another observation identifies where the gain comes from. OPD+TD changes two
channels of the student's view simultaneously: The \emph{computation graph} determines whether the student's forward pass actually hides the removed spans from later positions. The \emph{loss view} specifies the context under which the per-token training signal is computed. The paired design allows the two
channels to be altered independently at matched norm ($1.0\times$): masking
only the attention channel ($-0.004$, 95\% CI $[-0.005, -0.003]$,
$p<0.001$) or only the loss-view channel ($-0.002$, 95\% CI $[-0.0028,
-0.0003]$, $p=0.013$) performs \emph{worse} than the full-context OPD
update, whereas altering both jointly --- the full OPD+TD view --- yields the
$+0.027$ gain (Figure~\ref{fig:direction-schematic}b). The benefit therefore
does not arise from context masking per se, e.g.\ as regularization or noise
injection; it requires the coupled view change that defines OPD+TD, in which
the training signal and the computation graph are altered consistently.

In summary, on a model trained with standard OPD and under the condition of
identical update norms, the OPD+TD update is still better than the standard
OPD update: its advantage is directional and cannot be explained by
differences in gradient or update magnitude
(Figure~\ref{fig:direction-schematic}).

Paired analysis on fixed trajectories shows that TD strengthens the otherwise weak training signals in the PISA diagnostic regions. The norm-matched update experiment further demonstrates that TD produces updates that more effectively improve teacher alignment on held-out validation data. Together with the accuracy gains under standard full-context inference, these results support the proposed mechanism that TD improves learning by alleviating prefix-induced supervision attenuation.  

%% file: Appendix/appendix.tex
\section{Analysis Details}

\subsection{Scope of the sampled OPD gradient}
\label{app:opd}

We use the notation of Sections~\ref{sec:opd} and~\ref{sec:mechanism}.
All identities in this subsection are evaluated on a fixed rollout batch at
$\theta=\bar\theta$, before optimizer transformations.

For the full-context surrogate in Equation~\ref{eq:opd-loss},
$q_{\theta,t}/q_t=1$ at $\theta=\bar\theta$, and
\[
\left.
\nabla_\theta\frac{q_{\theta,t}}{q_t}
\right|_{\bar\theta}
=s_t^H.
\]
Since the advantage is detached, differentiating
Equation~\ref{eq:opd-loss} gives
\begin{equation}
 \nabla_\theta\mathcal L_{\mathrm{OPD}}(\bar\theta)
 =-\frac1N\sum_t A_t^H s_t^H.
\end{equation}
Thus $u_t^H$ in Equation~\ref{eq:opd-product} is the negative
per-token contribution to the loss gradient at the start of the update.

For a fixed prefix $h$, consider
$J_h(\theta)=D_{\mathrm{KL}}(Q_\theta(\cdot\mid h)\Vert
P_T(\cdot\mid h))$ and define
$A_\theta^H(v)=\log P_T(v\mid h)-\log Q_\theta(v\mid h)$.
Using $\sum_v Q_\theta(v\mid h)=1$,
\begin{align}
 \nabla_\theta J_h(\theta)
 &=\sum_v Q_\theta(v\mid h)
 \left(
 \log\frac{Q_\theta(v\mid h)}{P_T(v\mid h)}+1
 \right)
 \nabla_\theta\log Q_\theta(v\mid h)\\
 &=-\mathbb E_{v\sim Q_\theta(\cdot\mid h)}
 \left[
 A_\theta^H(v)\nabla_\theta\log Q_\theta(v\mid h)
 \right].
\end{align}
This identity conditions on a fixed prefix. Likewise, our sampled-token
analysis conditions on prefixes generated by $\bar\theta$ and therefore
characterizes the fixed-batch update rather than derivatives through a
$\theta$-dependent rollout distribution.

\subsection{Attention mask and span sampling}
\label{app:mask}

Let the prompt length be $P$.
Response token $s$ has key/value (K/V) index $P+s$, and the query for
target $t$ has index $P+t-1$.
For a set of selected spans $\mathcal S(M)$, the attention visibility mask is
\begin{equation}
 V_{ij}^M=\mathbf1[j\leq i]
 \prod_{[a,b)\in\mathcal S(M)}
 \left(
 1-\mathbf1[P+a\leq j<P+b]\mathbf1[i\geq P+b-1]
 \right).
 \label{eq:visibility}
\end{equation}
The same rule is applied before attention softmax in every layer and head.
All source positions and target losses remain in the computation; only the
selected direct attention edges are removed. 

For $T>1$ and $\rho>0$, the number of selected source positions is
\begin{equation}
 B=\min\{T-1,\max(1,\lfloor\rho T+0.5\rfloor)\}.
\end{equation}
Eligible sources exclude the final response position.
The sampler draws
$K$ uniformly from
$1,\ldots,\min(K_{\max},B,T-B)$,
samples positive span widths summing to $B$, and distributes the remaining
source positions across the gaps, with at least one visible position between
consecutive spans.
We use $K_{\max}=8$.
For $\rho=0$ or $T\leq1$, $\mathcal S(M)$ is empty.
Because $\rho$ measures selected source positions rather than removed
attention edges, the fraction of removed edges also depends on span locations.

\subsection{Confidence saturation and the correction curve}
\label{app:confidence}

Let $V$ be the vocabulary size, $\mathbf q$ the student probability vector,
$z$ its logits, and $q$ the probability of the sampled token.
The softmax identity gives
\begin{equation}
 \nabla_z\log q=e_y-\mathbf q,\qquad
 \|e_y-\mathbf q\|_2^2
 =(1-q)^2+\sum_{v\ne y}q_v^2.
 \label{eq:softmax-bound}
\end{equation}
Since the remaining probabilities sum to $1-q$,
Cauchy--Schwarz gives
\begin{equation}
 \frac{V}{V-1}(1-q)^2
 \leq
 \|e_y-\mathbf q\|_2^2
 \leq
 2(1-q)^2.
\end{equation}

Let $J=\partial z/\partial\theta$ be the logit Jacobian.
The parameter score is $s=J^\top(e_y-\mathbf q)$.
If $\|J\|_2\leq K$, then
\begin{equation}
 \|u^H\|_2
 =|A^H|\|J^\top(e_y-\mathbf q)\|_2
 \leq
 \sqrt{2}K|A^H|(1-q).
 \label{eq:parameter-bound}
\end{equation}
For fixed teacher probability $p>0$, $|A^H|=|\log(p/q)|$ remains bounded
as $q\to1$, so the parameter-level update also vanishes in this limit.

We next characterize the correction range used in
Equation~\ref{eq:confidence-gain}.
For fixed $p\in(0,1)$, define
\[
 F_p(v)=(1-v)\log(v/p),\qquad v\in(p,1).
\]
Its derivatives are
\begin{equation}
 F_p'(v)=v^{-1}-1-\log(v/p),\qquad
 F_p''(v)=-v^{-2}-v^{-1}<0.
\end{equation}
Thus $F_p$ is strictly concave.
Moreover,
$F_p'(p)=p^{-1}-1>0$ and $F_p'(1)=\log p<0$,
so it has a unique maximizer satisfying
$v^{-1}-1=\log(v/p)$.
Equivalently,
\begin{equation}
 v_p^\star=\frac1{W(e/p)},
\end{equation}
where $W$ is the positive Lambert function.
Hence $F_p$ decreases on $[v_p^\star,1)$, and
\[
 v_p^\star\leq r<q<1
 \quad\Longrightarrow\quad
 F_p(r)>F_p(q).
\]
This result concerns the sampled token's own-logit correction;
the full parameter update additionally depends on the remaining logits and
the network Jacobian.

\subsection{Agreement and update decomposition}
\label{app:agreement}

From Equation~\ref{eq:opd-product},
\begin{equation}
 |A^H|\leq\varepsilon_A
 \quad\Longrightarrow\quad
 \|u^H\|_2\leq\varepsilon_A\|s^H\|_2.
 \label{eq:agreement}
\end{equation}

For fixed teacher scores,
Equation~\ref{eq:adv-decomposition} gives
\[
 A^M-A^H=\log q-\log r=\delta.
\]
Adding and subtracting $A^Hs^M$ yields
Equation~\ref{eq:update-decomposition}.
Under exact agreement,
\begin{equation}
 A^H=0
 \quad\Longrightarrow\quad
 u^M=\delta s^M.
 \label{eq:agreement-recovery}
\end{equation}

More generally, when $|A^H|\leq\varepsilon_A$,
the reverse triangle inequality gives
\[
 |A^M|
 =|A^H+\delta|
 \geq
 (|\delta|-\varepsilon_A)_+.
\]
Therefore,
\begin{equation}
 |A^M-\delta|\leq\varepsilon_A,\qquad
 \|u^M\|_2
 \geq
 (|\delta|-\varepsilon_A)_+\|s^M\|_2,
\end{equation}
and
\begin{equation}
 \|u^M-u^H\|
 \leq
 |\delta|\|s^M\|
 +\varepsilon_A\|s^M-s^H\|.
\end{equation}
Thus near-agreement positions can acquire a non-negligible masked-view
update through either the induced advantage $\delta$ or a change in the
score vector.

\subsection{Advantage Clipping and the Local Regression Interpretation}
\label{app:local_regression}

At $\theta=\bar{\theta}$, the masked ratio
$w_t^M(\theta)=r_{\theta,t}/r_{\bar{\theta},t}$
equals one. Differentiating the TD surrogate in Equation~\eqref{eq:td-loss}
therefore gives
\begin{equation}
\nabla_\theta \mathcal{L}_{\mathrm{TD}}(\bar{\theta})
=
-\frac{1}{N}
\sum_t
A_t^M s_t^M ,
\tag{23}
\end{equation}
where
$A_t^M=\log p_t-\log r_t$
and
$s_t^M=\nabla_\theta \log r_{\theta,t}|_{\bar{\theta}}$.

For fixed trajectories, teacher scores, and attention masks, define
\begin{equation}
\mathcal{Q}_M(\theta)
=
\frac{1}{2N}
\sum_t
\left(
\log r_{\theta,t}-\log p_t
\right)^2 .
\tag{24}
\end{equation}
Direct differentiation gives
\begin{equation}
\begin{aligned}
\nabla_\theta \mathcal{Q}_M(\bar{\theta})
&=
\frac{1}{N}
\sum_t
(\log r_t-\log p_t)s_t^M \\
&=
-\frac{1}{N}
\sum_t
A_t^M s_t^M \\
&=
\nabla_\theta
\mathcal{L}_{\mathrm{TD}}(\bar{\theta}) .
\end{aligned}
\tag{25}
\end{equation}
Thus, at the start of an update, the raw TD update admits a local
interpretation as squared regression of the masked student's
log-probability toward the teacher log-probability. This interpretation is
local to the fixed rollout batch and fixed attention mask.

\paragraph{Advantage clipping.}
To allow finite clipping of the detached token-level log-probability gap,
define
\begin{equation}
\psi_c(a)
=
\min\{c,\max\{-c,a\}\},
\qquad
a_{t,c}^M(\theta)
=
\operatorname{sg}
\left[
\psi_c
\left(
\log p_t-\log r_{\theta,t}
\right)
\right],
\tag{26}
\end{equation}
with $\psi_\infty(a)=a$.
The corresponding advantage-clipped TD surrogate is
\begin{equation}
\mathcal{L}_{\mathrm{TD},c}(\theta)
=
-\frac{1}{N}
\sum_t
w_t^M(\theta)
a_{t,c}^M(\theta).
\tag{27}
\end{equation}
Since $w_t^M(\bar{\theta})=1$, its gradient at the update start is
\begin{equation}
\nabla_\theta
\mathcal{L}_{\mathrm{TD},c}(\bar{\theta})
=
-\frac{1}{N}
\sum_t
\psi_c(A_t^M)s_t^M .
\tag{28}
\end{equation}
Setting $c=\infty$ recovers Equation~(23).

For finite $c$, define the Huber function
\begin{equation}
H_c(e)
=
\begin{cases}
\dfrac{1}{2}e^2, & |e|\le c, \\[4pt]
c|e|-\dfrac{1}{2}c^2, & |e|>c .
\end{cases}
\tag{29}
\end{equation}
Since $H_c'(e)=\psi_c(e)$ and $\psi_c$ is an odd function, the objective
\begin{equation}
\begin{aligned}
\mathcal{Q}_{M,c}(\theta)
&=
\frac{1}{N}
\sum_t
H_c
\left(
\log r_{\theta,t}-\log p_t
\right), \\
\nabla_\theta
\mathcal{Q}_{M,c}(\bar{\theta})
&=
-\frac{1}{N}
\sum_t
\psi_c(A_t^M)s_t^M
=
\nabla_\theta
\mathcal{L}_{\mathrm{TD},c}(\bar{\theta}) .
\end{aligned}
\tag{30}
\end{equation}
Therefore, the local regression interpretation extends from squared loss
to Huber regression when the detached advantage is clipped.

\paragraph{Residual energy and advantage-clipped updates.}
Recall from Equation~\ref{eq:adv-decomposition} that
$A_t^M=A_t^H+\delta_t$, where
$\delta_t=\log q_t-\log r_t$.
The change in raw squared residual energy is
\begin{equation}
\frac{1}{2}(A_t^M)^2
-
\frac{1}{2}(A_t^H)^2
=
A_t^H\delta_t
+
\frac{1}{2}\delta_t^2 .
\tag{31}
\end{equation}
Under exact full-view agreement, $A_t^H=0$, so the masked view introduces
an additional squared residual of $\delta_t^2/2$.
This quantity characterizes the additional local regression residual
created by the prefix intervention; its effect on a full-context
validation objective still depends on the resulting parameter-update
direction and the optimizer.

For the advantage-clipped update, define the token-level contributions
\[
u_{t,c}^H
=
\psi_c(A_t^H)s_t^H,
\qquad
u_{t,c}^M
=
\psi_c(A_t^M)s_t^M .
\]
Using $A_t^M=A_t^H+\delta_t$, their difference can be decomposed as
\begin{equation}
\begin{aligned}
u_{t,c}^M-u_{t,c}^H
&=
\psi_c(A_t^H)
\left(
s_t^M-s_t^H
\right) \\
&\quad+
\left[
\psi_c(A_t^H+\delta_t)
-
\psi_c(A_t^H)
\right]
s_t^M .
\end{aligned}
\tag{32}
\end{equation}
The first term captures the change in the student's parameter sensitivity
under the masked view, while the second captures the change in the
advantage induced by altered prefix access. Thus, advantage clipping
preserves the same two-channel decomposition as the raw update, while
bounding the contribution of large token-level log-probability gaps.

\subsection{Sign Consistency of Confidence-Saturated Corrections}
\label{app:sign-reversal}

\paragraph{Sign reversals under prefix masking.}
Equation~(7) gives a sufficient condition under which Trajectory Dropout strengthens
the downward correction of a confidence-saturated token. This condition does not
hold for every sampled mask. In particular, consider a token satisfying
$p_t < q_t$, so that its full-view advantage is negative,
$A_t^H = \log p_t - \log q_t < 0$.
If prefix masking reduces the sampled-token probability below the teacher probability,
i.e., $r_t < p_t$, then
$A_t^M = \log p_t - \log r_t > 0,$
and the masked-view sampled-token advantage reverses sign.

Such reversals do occur in our diagnostic analysis, but they are rare and contribute
negligibly to the overall training signal. Among confidence-saturated tokens, only
approximately $8\%$ exhibit an advantage sign reversal under the masked view.
Moreover, these reversed tokens account for less than $5\%$ of the total signal
mass across all tokens. Thus, while sign reversal is possible outside the sufficient
regime characterized by Equation~(7), it affects only a small subset of the
confidence-saturated region and represents a negligible fraction of the global
training signal.

We emphasize that this is a reversal of the sampled-token advantage, and consequently
of its own-logit correction, rather than a statement that the full parameter-gradient
vector is reversed, since the score vector also changes from $s_t^H$ to $s_t^M$.

\paragraph{A sign-consistent safeguard.}
If strict sign consistency is desired, a simple safeguard is to fall back to the
standard OPD view whenever masking reverses the sign of the sampled-token advantage.
Define
$g_t =
\mathbf{1}\!\left[A_t^H A_t^M \ge 0\right].$
The corresponding sign-consistent update is
\begin{equation}
u_t^{\mathrm{SC}}
=
g_t A_t^M s_t^M
+
(1-g_t)A_t^H s_t^H.
\tag{33}
\end{equation}
Thus, TD uses the masked-view update when the advantage sign is preserved and
recovers the original full-context OPD update otherwise. The gate is determined
before the parameter update and is treated as fixed during optimization.
Importantly, the fallback restores both the student computation view and the
corresponding training signal to the full-context OPD view, rather than combining
a full-view advantage with a masked-view score vector. This preserves the
view consistency studied in Appendix~\ref{app:norm-controlled}.

%% file: iclr2027_conference.bib
@inproceedings{gu2024minillm,
 author = {Gu, Yuxian and Dong, Li and Wei, Furu and Huang, Minlie},
 booktitle = {International Conference on Learning Representations},
 editor = {B. Kim and Y. Yue and S. Chaudhuri and K. Fragkiadaki and M. Khan and Y. Sun},
 pages = {32694--32717},
 title = {MiniLLM: Knowledge Distillation of Large Language Models},
 url = {https://proceedings.iclr.cc/paper_files/paper/2024/file/8ac015d409635f196f9e3e9dcfb9a94e-Paper-Conference.pdf},
 volume = {2024},
 year = {2024}
}

@misc{agarwal2024gkd,
      title={On-Policy Distillation of Language Models: Learning from Self-Generated Mistakes}, 
      author={Rishabh Agarwal and Nino Vieillard and Yongchao Zhou and Piotr Stanczyk and Sabela Ramos and Matthieu Geist and Olivier Bachem},
      year={2024},
      eprint={2306.13649},
      archivePrefix={arXiv},
      primaryClass={cs.LG},
      url={https://arxiv.org/abs/2306.13649}, 
}

@inproceedings{fastopd2026,
    title = "Fast and Effective On-Policy Distillation from Reasoning Prefixes",
    author = "Zhang, Dongxu  and
      Yang, Zhichao  and
      Janghorbani, Sepehr  and
      Han, Jun  and
      II, Andrew Ressler  and
      Qian, Qian  and
      Lyng, Gregory D  and
      Batra, Sanjit Singh  and
      Tillman, Robert E.",
    editor = "Liakata, Maria  and
      Moreira, Viviane P.  and
      Zhang, Jiajun  and
      Jurgens, David",
    booktitle = "Findings of the {A}ssociation for {C}omputational {L}inguistics: {ACL} 2026",
    month = jul,
    year = "2026",
    address = "San Diego, California, United States",
    publisher = "Association for Computational Linguistics",
    url = "https://aclanthology.org/2026.findings-acl.1276/",
    doi = "10.18653/v1/2026.findings-acl.1276",
    pages = "25553--25569",
    ISBN = "979-8-89176-395-1"
}

@misc{opsd2026,
      title={Self-Distilled Reasoner: On-Policy Self-Distillation for Large Language Models}, 
      author={Siyan Zhao and Zhihui Xie and Mengchen Liu and Jing Huang and Guan Pang and Feiyu Chen and Aditya Grover},
      year={2026},
      eprint={2601.18734},
      archivePrefix={arXiv},
      primaryClass={cs.LG},
      url={https://arxiv.org/abs/2601.18734}, 
}

@misc{opcd2026,
      title={On-Policy Context Distillation for Language Models}, 
      author={Tianzhu Ye and Li Dong and Xun Wu and Shaohan Huang and Furu Wei},
      year={2026},
      eprint={2602.12275},
      archivePrefix={arXiv},
      primaryClass={cs.CL},
      url={https://arxiv.org/abs/2602.12275}, 
}

@misc{s2vopd2026,
      title={Self-Supervised Visual On-Policy Distillation}, 
      author={Yijiang Li and Yijun Liang and Yunjie Tian and Bingyang Wang and Ke Zhang and Zhenfei Yin and Di Fu and Philip Torr and Nuno Vasconcelos},
      year={2026},
      eprint={2608.14144},
      archivePrefix={arXiv},
      primaryClass={cs.CV},
      url={https://arxiv.org/abs/2608.14144}, 
}

@inproceedings{reflectioncoder2025,
    title = "{R}eflection{C}oder: Learning from Reflection Sequence for Enhanced One-off Code Generation",
    author = "Ren, Houxing  and
      Zhan, Mingjie  and
      Wu, Zhongyuan  and
      Zhou, Aojun  and
      Pan, Junting  and
      Li, Hongsheng",
    editor = "Che, Wanxiang  and
      Nabende, Joyce  and
      Shutova, Ekaterina  and
      Pilehvar, Mohammad Taher",
    booktitle = "Proceedings of the 63rd Annual Meeting of the Association for Computational Linguistics (Volume 1: Long Papers)",
    month = jul,
    year = "2025",
    address = "Vienna, Austria",
    publisher = "Association for Computational Linguistics",
    url = "https://aclanthology.org/2025.acl-long.494/",
    doi = "10.18653/v1/2025.acl-long.494",
    pages = "9999--10020",
    ISBN = "979-8-89176-251-0"
}

@misc{hidetosee2026,
      title={Hide to See: Reasoning-prefix Masking for Visual-anchored Thinking in VLM Distillation}, 
      author={Seonghoon Yu and Dongjun Nam and Byung-Kwan Lee and Jeany Son},
      year={2026},
      eprint={2605.11651},
      archivePrefix={arXiv},
      primaryClass={cs.CV},
      url={https://arxiv.org/abs/2605.11651}, 
}

@misc{adaptivefastopd2026,
      title={Adaptive FastOPD: Progress-Aware Rollout Horizon Expansion for Efficient On-Policy Distillation}, 
      author={Qian Tan and Huaifei Liang and Xuanyu Zhu and Lei Jiang and Yuqiang Li},
      year={2026},
      eprint={2607.29494},
      archivePrefix={arXiv},
      primaryClass={cs.LG},
      url={https://arxiv.org/abs/2607.29494}, 
}

@misc{tip2026,
      title={TIP: Token Importance in On-Policy Distillation}, 
      author={Yuanda Xu and Hejian Sang and Zhengze Zhou and Ran He and Zhipeng Wang and Alborz Geramifard},
      year={2026},
      eprint={2604.14084},
      archivePrefix={arXiv},
      primaryClass={cs.LG},
      url={https://arxiv.org/abs/2604.14084}, 
}

@misc{fireopd2026,
      title={Filter, Then Reweight: Rethinking Optimization Granularity in On-Policy Distillation}, 
      author={Yuying Li and Leqi Zheng and Yongzi Yu and Wenrui Zhou and Xuchang Zhong and Xing Hu and Jing Jin and Hangjie Yuan and Tao Feng},
      year={2026},
      eprint={2606.02684},
      archivePrefix={arXiv},
      primaryClass={cs.LG},
      url={https://arxiv.org/abs/2606.02684}, 
}

@misc{taopd2026,
      title={Not All Disagreement Is Learnable: Token Teachability in On-Policy Distillation}, 
      author={Yuanyi Wang and Su Lu and Yanggan Gu and Pengkai Wang and Yifan Yang and Zhaoyi Yan and Congkai Xie and Jianmin Wu and Hongxia Yang},
      year={2026},
      eprint={2605.26844},
      archivePrefix={arXiv},
      primaryClass={cs.LG},
      url={https://arxiv.org/abs/2605.26844}, 
}

@misc{prefixteach2026,
      title={Prefix Teach, Suffix Fade: Local Teachability Collapse in Strong-to-Weak On-Policy Distillation}, 
      author={Kaiyuan Liu and Ziyuan Zhuang and Yang Bai and Bing Wang and Rongxiang Weng and Jieping Ye},
      year={2026},
      eprint={2605.13643},
      archivePrefix={arXiv},
      primaryClass={cs.CL},
      url={https://arxiv.org/abs/2605.13643}, 
}

@misc{renio2026,
      title={ReNIO: Reweighting Negative Trajectory Importance for LLM On-Policy Distillation}, 
      author={Chen Lin and Kedi Chen and Wei Zhang},
      year={2026},
      eprint={2606.23104},
      archivePrefix={arXiv},
      primaryClass={cs.LG},
      url={https://arxiv.org/abs/2606.23104}, 
}

@misc{trd2026,
      title={Trajectory-Refined Distillation}, 
      author={Li Jiang and Haoran Xu and Yichuan Ding and Amy Zhang},
      year={2026},
      eprint={2606.08432},
      archivePrefix={arXiv},
      primaryClass={cs.AI},
      url={https://arxiv.org/abs/2606.08432}, 
}

@misc{topd2026,
      title={Bridging Reasoning Trajectories in On-Policy Distillation via Near-Future Guidance}, 
      author={Yuxuan Jiang and Francis Ferraro},
      year={2026},
      eprint={2606.00305},
      archivePrefix={arXiv},
      primaryClass={cs.CL},
      url={https://arxiv.org/abs/2606.00305}, 
}

@misc{nopd2026,
      title={Self-Boosting Vision-Language Models with Noisy Student On-Policy Self-Distillation}, 
      author={Shuai Wang and Daoan Zhang and Zhe Tang and Hao Cheng and Jiaheng Wei},
      year={2026},
      eprint={2607.23125},
      archivePrefix={arXiv},
      primaryClass={cs.LG},
      url={https://arxiv.org/abs/2607.23125}, 
}

@misc{gao2025tokendropout,
      title={What Makes Diffusion Language Models Super Data Learners?}, 
      author={Zitian Gao and Haoming Luo and Lynx Chen and Jason Klein Liu and Ran Tao and Joey Zhou and Bryan Dai},
      year={2025},
      eprint={2510.04071},
      archivePrefix={arXiv},
      primaryClass={cs.CL},
      url={https://arxiv.org/abs/2510.04071}, 
}

@misc{chen2026pretrainaug,
      title={Demystifying Training-Time Augmentation for Data-Constrained Language Model Pretraining}, 
      author={Michael K. Chen and Xikun Zhang and Fan Bai and Zhengding Hu and Zhen Wang},
      year={2026},
      eprint={2606.16246},
      archivePrefix={arXiv},
      primaryClass={cs.LG},
      url={https://arxiv.org/abs/2606.16246}, 
}

@inproceedings{
trsd2026,
title={Learning from Partial Chain-of-Thought via Truncated-Reasoning Self-Distillation},
author={Gianluigi Silvestri and Edoardo Cetin},
booktitle={Workshop on Latent {\&} Implicit Thinking {\textendash} Going Beyond CoT Reasoning},
year={2026},
url={https://openreview.net/forum?id=HNXlTFIGsl}
}

@inproceedings{
maskeddistillation2026,
title={Masked Distillation: Internalizing the Chain-of-Thought in Language Models},
author={Durgesh Kalwar and Vardhan Palod and Subbarao Kambhampati},
booktitle={ICML 2026 Workshop on Foundations of Deep Generative Models: Understanding Memorization, Generalization, and Reasoning},
year={2026},
url={https://openreview.net/forum?id=Zt36ZCqZWU}
}

@misc{liu2026fidelity,
      title={Your Teacher Can't Help You Here: Combating Supervision Fidelity Decay in On-Policy Distillation}, 
      author={Yanjiang Liu and Jie Lou and Xinyan Guan and Yuqiu Ji and Hongyu Lin and Ben He and Xianpei Han and Le Sun and Xing Yu and Yaojie Lu},
      year={2026},
      eprint={2605.30833},
      archivePrefix={arXiv},
      primaryClass={cs.CL},
      url={https://arxiv.org/abs/2605.30833}, 
}

@misc{armandpour2026unmasking,
      title={Unmasking On-Policy Distillation: Where It Helps, Where It Hurts, and Why}, 
      author={Mohammadreza Armandpour and Fatih Ilhan and David Harrison and Ajay Jaiswal and Duc N. M Hoang and Fartash Faghri and Yizhe Zhang and Minsik Cho and Mehrdad Farajtabar},
      year={2026},
      eprint={2605.10889},
      archivePrefix={arXiv},
      primaryClass={cs.LG},
      url={https://arxiv.org/abs/2605.10889}, 
}

@inproceedings{he2024olympiadbench,
    title = "{O}lympiad{B}ench: A Challenging Benchmark for Promoting {AGI} with Olympiad-Level Bilingual Multimodal Scientific Problems",
    author = "He, Chaoqun  and
      Luo, Renjie  and
      Bai, Yuzhuo  and
      Hu, Shengding  and
      Thai, Zhen  and
      Shen, Junhao  and
      Hu, Jinyi  and
      Han, Xu  and
      Huang, Yujie  and
      Zhang, Yuxiang  and
      Liu, Jie  and
      Qi, Lei  and
      Liu, Zhiyuan  and
      Sun, Maosong",
    editor = "Ku, Lun-Wei  and
      Martins, Andre  and
      Srikumar, Vivek",
    booktitle = "Proceedings of the 62nd Annual Meeting of the Association for Computational Linguistics (Volume 1: Long Papers)",
    month = aug,
    year = "2024",
    address = "Bangkok, Thailand",
    publisher = "Association for Computational Linguistics",
    url = "https://aclanthology.org/2024.acl-long.211/",
    doi = "10.18653/v1/2024.acl-long.211",
    pages = "3828--3850"
}

@misc{yang2025qwen3,
      title={Qwen3 Technical Report}, 
      author={An Yang and Anfeng Li and Baosong Yang and Beichen Zhang and Binyuan Hui and Bo Zheng and Bowen Yu and Chang Gao and Chengen Huang and Chenxu Lv and Chujie Zheng and Dayiheng Liu and Fan Zhou and Fei Huang and Feng Hu and Hao Ge and Haoran Wei and Huan Lin and Jialong Tang and Jian Yang and Jianhong Tu and Jianwei Zhang and Jianxin Yang and Jiaxi Yang and Jing Zhou and Jingren Zhou and Junyang Lin and Kai Dang and Keqin Bao and Kexin Yang and Le Yu and Lianghao Deng and Mei Li and Mingfeng Xue and Mingze Li and Pei Zhang and Peng Wang and Qin Zhu and Rui Men and Ruize Gao and Shixuan Liu and Shuang Luo and Tianhao Li and Tianyi Tang and Wenbiao Yin and Xingzhang Ren and Xinyu Wang and Xinyu Zhang and Xuancheng Ren and Yang Fan and Yang Su and Yichang Zhang and Yinger Zhang and Yu Wan and Yuqiong Liu and Zekun Wang and Zeyu Cui and Zhenru Zhang and Zhipeng Zhou and Zihan Qiu},
      year={2025},
      eprint={2505.09388},
      archivePrefix={arXiv},
      primaryClass={cs.CL},
      url={https://arxiv.org/abs/2505.09388}, 
}

@inproceedings{
yu2025dapo,
title={{DAPO}: An Open-Source {LLM} Reinforcement Learning System at Scale},
author={Qiying Yu and Zheng Zhang and Ruofei Zhu and Yufeng Yuan and Xiaochen Zuo and YuYue and Weinan Dai and Tiantian Fan and Gaohong Liu and Juncai Liu and LingJun Liu and Xin Liu and Haibin Lin and Zhiqi Lin and Bole Ma and Guangming Sheng and Yuxuan Tong and Chi Zhang and Mofan Zhang and Ru Zhang and Wang Zhang and Hang Zhu and Jinhua Zhu and Jiaze Chen and Jiangjie Chen and Chengyi Wang and Hongli Yu and Yuxuan Song and Xiangpeng Wei and Hao Zhou and Jingjing Liu and Wei-Ying Ma and Ya-Qin Zhang and Lin Yan and Yonghui Wu and Mingxuan Wang},
booktitle={The Thirty-ninth Annual Conference on Neural Information Processing Systems},
year={2025},
url={https://openreview.net/forum?id=2a36EMSSTp}
}

@inproceedings{helm-etal-2025-token,
    title = "Token Weighting for Long-Range Language Modeling",
    author = "Helm, Falko  and
      Daheim, Nico  and
      Gurevych, Iryna",
    editor = "Chiruzzo, Luis  and
      Ritter, Alan  and
      Wang, Lu",
    booktitle = "Findings of the Association for Computational Linguistics: NAACL 2025",
    month = apr,
    year = "2025",
    address = "Albuquerque, New Mexico",
    publisher = "Association for Computational Linguistics",
    url = "https://aclanthology.org/2025.findings-naacl.79/",
    doi = "10.18653/v1/2025.findings-naacl.79",
    pages = "1440--1459",
    ISBN = "979-8-89176-195-7"
}

@inproceedings{
deng2026beyond,
title={Beyond Length: Quantifying Long-Range Information for Long-Context {LLM} Pretraining Data},
author={Haoran Deng and Yingyu Lin and Zhenghao Lin and Xiao Liu and Yizhou Sun and Yian Ma and Yeyun Gong},
booktitle={The Fourteenth International Conference on Learning Representations},
year={2026},
url={https://openreview.net/forum?id=C9TDQ8Wwx7}
}

@misc{lin2026ccopd,
      title={Same Evidence, Different Answers: Canonical-Context On-Policy Distillation for Multi-Turn Language Models}, 
      author={Zizhuo Lin and Quanling Liu and Jinsheng Quan and Chao Zhang and Yifan Zhu and Xing Shi and Jingtao Xu and Zhihui Li and Yawei Luo},
      year={2026},
      eprint={2605.30251},
      archivePrefix={arXiv},
      primaryClass={cs.CL},
      url={https://arxiv.org/abs/2605.30251}, 
}

@misc{gbqa,
      title={GPQA: A Graduate-Level Google-Proof Q\&A Benchmark}, 
      author={David Rein and Betty Li Hou and Asa Cooper Stickland and Jackson Petty and Richard Yuanzhe Pang and Julien Dirani and Julian Michael and Samuel R. Bowman},
      year={2023},
      eprint={2311.12022},
      archivePrefix={arXiv},
      primaryClass={cs.AI},
      url={https://arxiv.org/abs/2311.12022}, 
}

@inproceedings{
mbpp,
title={Is Your Code Generated by Chat{GPT} Really Correct? Rigorous Evaluation of Large Language Models for Code Generation},
author={Jiawei Liu and Chunqiu Steven Xia and Yuyao Wang and LINGMING ZHANG},
booktitle={Thirty-seventh Conference on Neural Information Processing Systems},
year={2023},
url={https://openreview.net/forum?id=1qvx610Cu7}
}

@misc{2026dense,
      title={Dense Supervision, Sparse Updates: On the Sparsity and Geometry of On-Policy Distillation}, 
      author={Guo Yu and Wenlin Liu and Yulan Hu and Hao-Xuan Ma and Jun-Peng Jiang and Han-Jia Ye},
      year={2026},
      eprint={2606.13657},
      archivePrefix={arXiv},
      primaryClass={cs.LG},
      url={https://arxiv.org/abs/2606.13657}, 
}

@misc{shao2026token,
      title={A Token-Level Analysis of Sampled-Token Reverse-KL On-Policy Distillation}, 
      author={Bing Shao and Jiazheng Zhang and Long Ma and Yujiong Shen and Senjie Jin and Xin Guo and Yuming Yang and Mingxu Chai and Zhiheng Xi and Boyang Liu and Junlin Shang and Tao Gui and Qi Zhang and Xuanjing Huang},
      year={2026},
      eprint={2608.25643},
      archivePrefix={arXiv},
      primaryClass={cs.LG},
      url={https://arxiv.org/abs/2608.25643}, 
}
